\documentclass[]{phai}

\usepackage{amssymb}
\usepackage{mathtools}
\usepackage{wasysym}
\usepackage{makecell}
\usepackage[normalem]{ulem}
\usepackage{stackengine}
\usepackage{enumitem}
\usepackage{array}
\usepackage{colortbl}
\usepackage{footmisc}
\usepackage{xspace}
\usepackage{xargs}
\usepackage{threeparttable}
\usepackage{adjustbox}

\providecolor{ForestGreen}{RGB}{34,139,34}

\newcommand{\best}[1]{\textbf{#1}}

\newcommand{\BenchmarkName}{{CrashTwin}}

\title{A Physics-Grounded Benchmark for \\Multi-Agent Dynamics in World Models}

\author[*1]{Nuo Chen}
\author[*1,2]{Lulin Liu}
\author[3]{Zihao Li}
\author[4]{Ziyao Zeng}
\author[1]{Zihao Zhu}
\author[5]{Wenyan Cong}
\author[6,7]{Junyuan Hong}
\author[5]{Yunhao Yang}
\author[1]{Zhengzhong Tu}
\author[8]{Yan Wang}
\author[8]{Boris Ivanovic}
\author[8,9]{Marco Pavone}
\author[5]{Zhangyang Wang}
\author[1]{Yang Zhou}
\author[1]{Zhiwen Fan}

\affiliation[1]{Texas A\&M University}
\affiliation[2]{University of Minnesota}
\affiliation[3]{Marquette University}
\affiliation[4]{Yale University}
\affiliation[5]{University of Texas at Austin}
\affiliation[6]{Massachusetts General Hospital}
\affiliation[7]{Harvard Medical School}
\affiliation[8]{NVIDIA}
\affiliation[9]{Stanford University}

\contribution[*]{Equal contribution}

\abstract{%
\begingroup
\renewenvironment{abstract}{}{}%

\begin{abstract}
Generative world models hold immense promise as scalable simulators for autonomous systems, particularly for synthesizing rare but safety-critical multi-agent interactions, such as vehicle collisions. However, current evaluation paradigms index heavily on visual fidelity and semantic alignment, leaving a critical blind spot: they cannot reliably quantify whether generated dynamics actually obey the fundamental physical laws required for reliable simulation. 
Assessing this physical plausibility is inherently difficult due to a lack of physical metrics and the challenge of extracting metric-scale kinematics from uncalibrated video rollouts.
To bridge this gap, we introduce \textbf{CrashTwin}, a physics-grounded evaluation framework designed to stress-test the physical trustworthiness of world models. 
CrashTwin couples a diverse dataset of multi-agent collision scenarios, comprising 25K controllable synthetic and 12K in-the-wild real-world collision sequences with a novel calibration-free reconstruction pipeline,
enabling the recovery of 3D physical attributes directly from world model rollouts. We propose a diagnostic suite that systematically evaluates three dimensions: spatio-temporal consistency, momentum and kinetic energy conservation, and world-dynamics integrity. Extensive benchmarking of state-of-the-art models reveals a crucial insight: high perceptual quality frequently masks severe physical violations during complex interactions. By quantitatively exposing these failure modes, CrashTwin provides a vital diagnostic tool for developing physically grounded world models capable of reliable real-world simulation.

\end{abstract}

\endgroup
}

\date{\today}
\metadata[Keywords]{World Models, Physics-Grounded Evaluation, Autonomous Driving}
\metadata[Project Website]{\url{https://crash-twin.github.io/}}

\begin{document}

\maketitle

\section{Introduction}
\label{sec:intro}

\begin{figure}[t]
    \centering
    \includegraphics[width=0.95\linewidth]{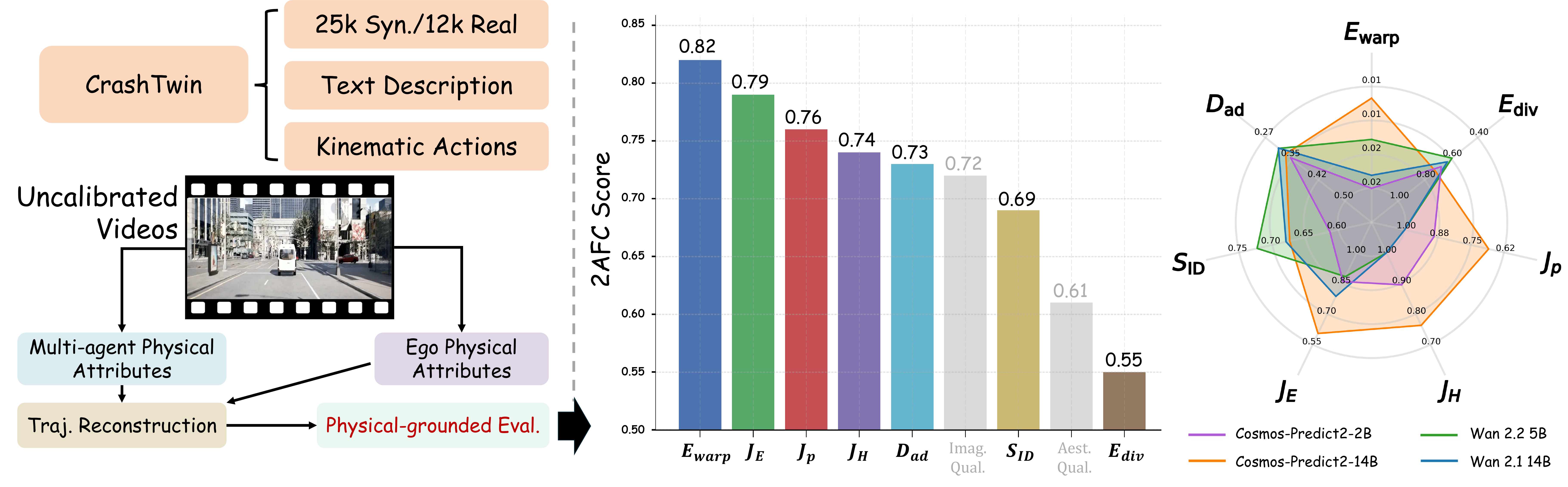}
\caption{\textbf{CrashTwin Benchmark.} \textbf{Left} illustrates the content of CrashTwin and the proposed evaluation pipeline that extracts physical attributes from uncalibrated videos to enable physics-grounded evaluation. \textbf{Middle} shows the two-alternative-forced-choice(2AFC Score), demonstrating that metrics derived from physical dynamics, indicated by colored bars, align more strongly with human preferences for physical realism compared to conventional visual quality proxies shown in grayscale. \textbf{Right} compares representative world models, revealing that both small and large models exhibit notable deficiencies across the proposed physics-grounded criteria.}
    \vspace{-6mm}
    \label{fig:orthogonality}
\end{figure}

Generative world models~\cite{Wang23ArXiv-DriveWM,Wen24CVPR-Panacea,Wang24ECCV-DriveDreamer,Hu23ArXiv-GAIA1,Zheng24ECCV-OccWorld} have emerged as promising tools for scalable simulation in autonomous driving, offering a pathway to synthesize safety-critical corner cases that are rare in the real world~\cite{waymo2021safetyreport, xu2025wod}. 
For these rollouts to be actionable in downstream data curation, they must not only exhibit high visual fidelity but also strictly adhere to physical laws. However, the current evaluation paradigm predominantly indexes on perceptual quality and semantic alignment~\cite{Huang24CVPR-VBench, zheng2025vbench2, Unterthiner18ArXiv-FVD}. Consequently, a critical blind spot remains: severe violations of fundamental physics frequently go undetected under standard visual assessments. Furthermore, these conventional visual metrics correlate poorly with human judgments of physical realism, as illustrated in~\cref{fig:orthogonality}.

Rigorously quantifying this physical plausibility is inherently difficult, particularly in safety-critical scenarios. 
Existing frameworks often circumvent the underlying mechanics by relying on generalized visual proxies, including distributional similarity metrics such as Fr{\'e}chet Video Distance~\cite{Unterthiner18ArXiv-FVD} and appearance-focused benchmark scores such as the aesthetic and imaging quality~\cite{Huang24CVPR-VBench,Unterthiner18ArXiv-FVD}, or vision-language models (VLMs) as judges~\cite{Bansal25ICLR-VideoPhy}, which remain brittle under visual ambiguity and are insufficient for precise physical validation~\cite{liu2025your, puyin2025quantiphy}.
The fundamental bottleneck lies in physical observability: to rigorously verify adherence to physical laws, one must extract metric-scale kinematics, such as precise trajectories, velocities, and angular motion. Yet, recovering these 3D physical attributes from uncalibrated monocular videos is a highly ill-posed problem, requiring the disentanglement of ego-motion, camera intrinsics, and multi-agent dynamics without explicit physical sensors. This difficulty is severely compounded during complex interactions featuring rapid relative motion and heavy occlusion. Furthermore, current video generation benchmarks primarily consist of general-domain or benign driving clips, lacking the targeted, safety-critical data required to expose these specific dynamic failures.

To address these methodological and data bottlenecks, we focus on vehicle-to-vehicle collisions as a representative task for evaluating multi-agent physical interactions. This setting provides a rigorous stress test for generative world models, as physical impacts induce abrupt, coupled state changes whose post-contact evolution is tightly governed by classical conservation laws~\cite{brach1987review, chatterjee1997rigid}. By focusing on these multi-agent interactions, we establish a mathematically grounded environment for quantitative analysis, yielding consistent geometric priors and clear kinematic constraints~\cite{zhou2008collision}. Because these specific scenarios are inherently physics-dependent, verifying their structural integrity is a crucial prerequisite before such world models can be trusted as reliable simulators.

In this work, we present \textbf{\BenchmarkName}, a comprehensive evaluation framework designed to quantify the physical integrity of world model rollouts. 
To ensure both rigorous control and real-world relevance, we construct an ever large dataset that contains comprehensive text descriptions, poses indicating agent positions and states, and kinematic action labels, built from 25K synthetic sequences generated via a physically grounded pipeline~\cite{Dosovitskiy17CoRL-CARLA} and 12K diverse in-the-wild traffic incident videos.
To overcome the challenge of evaluating uncalibrated world-model rollouts, we develop a calibration-free reconstruction pipeline that leverages vision foundation models to reliably recover physical attributes under complex interactions. 
Building upon these recovered 3D trajectories, we introduce a standardized diagnostic protocol evaluating three core dimensions: spatio-temporal consistency, momentum and energy conservation, and world-dynamics integrity. Extensive benchmarking demonstrates that our metrics expose physical failure modes largely invisible to appearance-based evaluations, offering an orthogonal diagnostic signal for developing physically trustworthy world models.

Our contributions are threefold:
\begin{itemize}
    \item We introduce \textbf{\BenchmarkName}, a large-scale benchmark coupling controllable synthetic data with unconstrained real-world collision videos, directly addressing the lack of safety-critical data in current evaluation paradigms.
    \item We develop a physics-grounded evaluation framework featuring a calibration-free reconstruction pipeline to recover metric-scale physical attributes from monocular videos, resolving the bottleneck of uncalibrated measurement to enable the precise evaluation of spatio-temporal consistency, momentum conservation, energy dissipation, and world dynamics.
    \item We systematically benchmark representative world models under this protocol, providing quantitative diagnoses of their physical failure modes and demonstrating that our metrics align significantly better with human preference judgments regarding physical realism.
\end{itemize}

\begin{figure*}[t]
    \centering
        
        \includegraphics[width=0.95\linewidth]{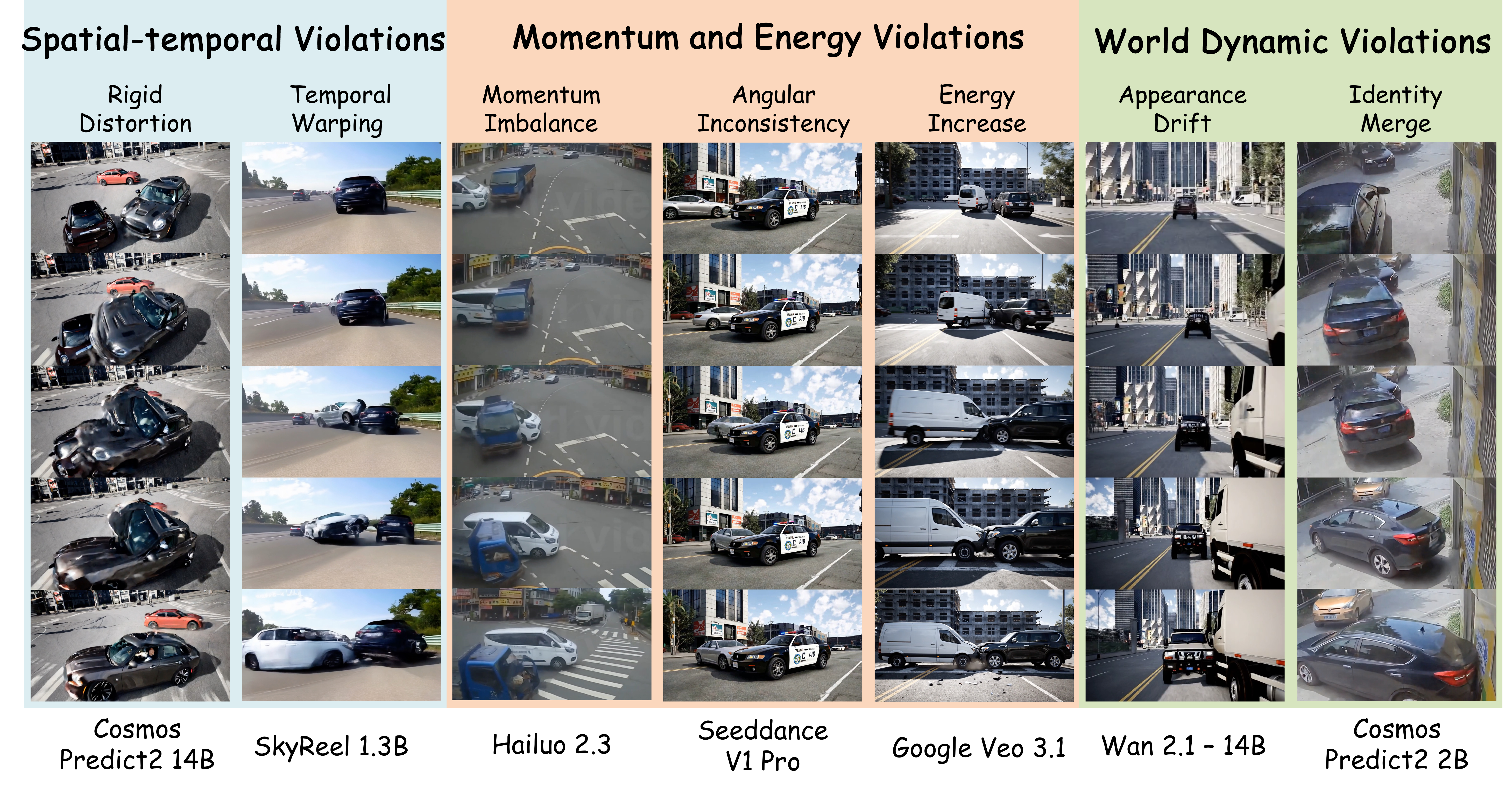}
\caption{\textbf{Failure cases of existing world models revealed by our benchmark. }Each region highlights a distinct type of physical breakdown, including violations of spatio-temporal consistency, departures from momentum and energy conservation around impact, and breakdowns of world-dynamics integrity such as unstable identities or appearances. 
These patterns illustrate the characteristic ways in which current methods deviate from physically consistent behavior.}
  \vspace{-6mm}  

    \label{fig:failcase}
\end{figure*}

\section{Related Works}

\paragraph{Video Generation and World Models.} Video generation has advanced rapidly with diffusion and autoregressive transformers, enabling high-fidelity text-to-video across diverse domains \cite{Han25CVPR-VideoBench,Sun24ArXiv-CompBench}. In autonomous driving, world models like GAIA-1, DriveDreamer-2, and Epona simulate controllable, multi-view traffic scenes with long-horizon rollouts, while egocentric simulators such as PlayerOne and GEM synthesize first-person videos from user motion or object controls \cite{Hu23ArXiv-GAIA1,Zhao24AAAI-DriveDreamer2,Zhang25ArXiv-Epona,Tu25ArXiv-PlayerOne,Hassan25CVPR-GEM,Huang24CVPR-VBench}. Yet appearance fidelity does not guarantee physical validity. Recent crash-oriented video generation models attempt to address safety-critical scenarios but fundamentally sidestep true physical simulation. 
For instance, DrivingGen \cite{guo2024drivinggen} is text-driven and does not fix the first-frame, preventing controlled comparisons of physical outcomes. Similarly, AVD2 \cite{li2025avd2} focuses on extreme ego-centric self-collisions where vital pre- and post-impact kinematic information is effectively destroyed, making their physical fidelity inherently untestable. 
Furthermore, methods like OAVD \cite{Fang_2024_CVPR} and Ctrl-Crash \cite{gosselin2025ctrl} strictly condition generation on comprehensive temporal bounding boxes. By providing the temporal object trajectories as input conditions, these models perform appearance completion within predefined constraints rather than actual physical simulation.
Because these specialized models either rely on strong physical priors or generate untestable ego-collisions, they lack the capacity for generalized physical reasoning. Consequently, benchmarking them for physical consistency is inherently meaningless, as the physics are either dictated by the input conditions or completely unrecoverable. Instead, evaluating the physical characteristics of unconstrained world models remains a critical open challenge. Studies show that current models often ignore kinematics and violate momentum during collisions \cite{Kang25ICML-PhysicalLaw,Bansal25ICLR-VideoPhy,Liu24ECCV-PhysGen}. This has motivated hybrids that embed classical rigid-body simulators and benchmarks that test whether generated trajectories obey fundamental laws rather than only look plausible \cite{Liu24ECCV-PhysGen,Bansal25ICLR-VideoPhy,Meng24ArXiv-PhyGenBench}. Our {\BenchmarkName} benchmark advances this direction by explicitly quantifying spatio temporal consistency, momentum and energy conservation, and instance level structural integrity during complex vehicle collisions.

\paragraph{Physical-grounded Evaluation.}
Most evaluation in world modeling is still perception-centric, prioritizing appearance fidelity, temporal coherence, and preference alignment over adherence to physical laws and dynamics. Distributional and perceptual scores such as FID, FVD, and embedding- or quality-based measures focus on visual realism instead of law adherence \cite{Heusel17NeurIPS-FID,Unterthiner18ArXiv-FVD}, and comprehensive benchmarks like VBench and DEVIL mainly decompose visible quality, including identity consistency, motion smoothness, temporal stability, and dynamics coherence, rather than testing mechanics \cite{Huang24CVPR-VBench,Liao24NeurIPS-DEVIL}. Recent “physics-aware” work follows three lines: (i) using LLMs or multimodal transformers as commonsense judges of plausibility \cite{Kang25ICML-PhysicalLaw,Bansal25ICLR-VideoPhy,Xue25CVPR-PhyT2V}; (ii) heuristic proxies enforcing optical-flow or multi-frame depth and pose consistency \cite{Wu24ArXiv-T2VScore,Zeng23ArXiv-EvalCrafter}; and (iii) scripted scenarios that flag violations or probe conservation in simplified collisions \cite{Meng24ArXiv-PhyGenBench,Kang25ICML-PhysicalLaw,Bansal25ICLR-VideoPhy}. These approaches are largely indirect or qualitative, avoid directly measuring physical quantities, are prompt- or design-sensitive, and rarely localize errors in space and time, which hampers reproducibility and objective comparison \cite{Huang24CVPR-VBench,Wu24ArXiv-T2VScore,Zeng23ArXiv-EvalCrafter,Meng24ArXiv-PhyGenBench}. The limitations are especially severe in collision scenes, where models still show ghosting through obstacles, momentum-inconsistent velocity changes, and invalid contact geometry even when appearance scores are high \cite{Bansal25ICLR-VideoPhy,Kang25ICML-PhysicalLaw}. This gap motivates physical-law-grounded verification with localized diagnosis. We introduce {\BenchmarkName }, which directly tests spatio-temporal consistency, momentum and energy conservation, and world-dynamics integrity in predicted collision events, providing quantitative localized checks that current metrics overlook and supporting progress toward deployable world models for planning and decision-making \cite{Zhang25ArXiv-Epona,Hafner19ICML-PlaNet}.

\section{Principles of the New Evaluation Protocol}

\label{sec:method}

\begin{figure*}[t]
    \centering
        \includegraphics[width=0.95\linewidth]{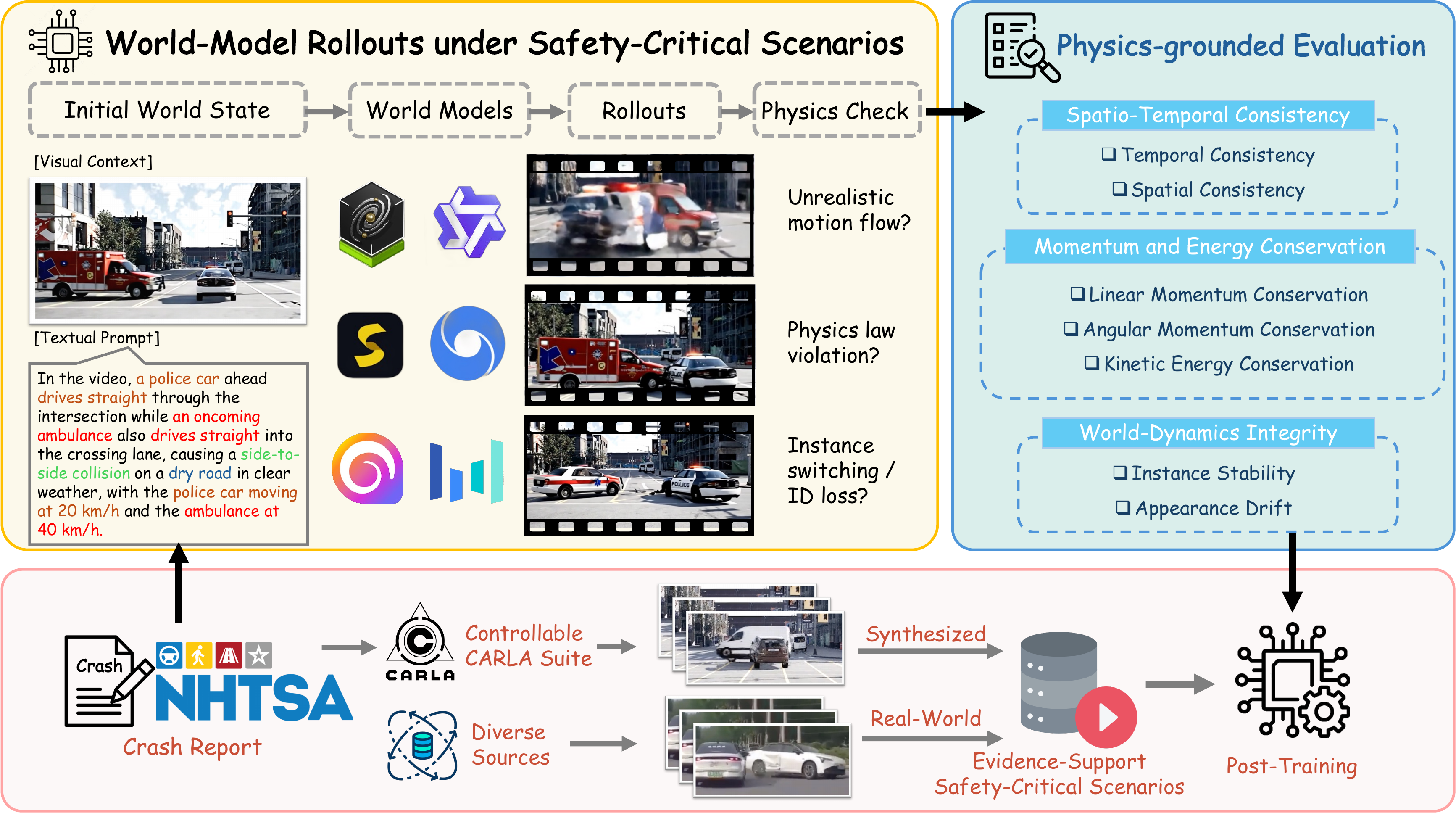}
    \caption{\textbf{Overview of the \BenchmarkName~framework.} We derive safety-critical collision from national pre-crash statistics and instantiate them in both a controllable CARLA suite and a diverse real-world corpus. Current world models often violate physical principles in these rollouts, motivating our physics-grounded evaluation framework that measures spatio-temporal consistency, momentum and energy conservation, and world-dynamics integrity. Guided by these metrics, post-training improves physical fidelity.}\label{fig:overall_pipeline}
    \vspace{-6mm}
    
\end{figure*}

The proposed \BenchmarkName~is a physics grounded framework for analyzing and evaluating crash events. Section~\ref{sec:phys} first presents three dimensions of physical behavior and their associated evaluation metrics. Unlike prior evaluations that focus on image, patch, or pixel level accuracy, \BenchmarkName~instead measures physical quantities and tests physical laws and contact constraints using measurable signals, localizing violations around first contact and reporting their magnitudes through physics-grounded metrics. Next, Section~\ref{sec:data} introduces a carefully controlled CARLA based data generation pipeline supplemented with a real world crash corpus. Finally, Section~\ref{sec:eval} details the calibration free reconstruction pipeline that recovers the necessary physical attributes from uncalibrated videos to support these evaluations.

\subsection{Physical-Grounded Evaluation}
\label{sec:phys}
We study physics-grounded behavior across three complementary dimensions: {spatio-temporal consistency}, {momentum and energy conservation}, and {world-dynamics integrity}, as illustrated in~\cref{fig:overall_pipeline}. Together, these dimensions provide a comprehensive measure of how well generated collisions adhere to physical laws, assessing whether motion remains continuous and bounded, impacts satisfy conservation principles, and interacting instances maintain stable, non-penetrating identities throughout the event.

\subsubsection{Spatio-Temporal Consistency}
We aim to evaluate whether generated motion is smooth and plausible, through \emph{temporal coherence}, ensuring continuous motion over time, and \emph{spatial rigidity}, requiring foreground instances keep consistent geometry and scale without unrealistic expansion or compression.

\paragraph{Determining Temporal Consistency.}
We measure temporal coherence using a \emph{flow warping error}, which quantifies how well pixels propagated by optical flow match their corresponding observations in the next frame:
\begin{equation}
E_{\text{warp}} = \frac{1}{|\Omega|}\sum_{p\in\Omega} \big| I_t(p) - I_{t+1}(p + F_{t\rightarrow t+1}(p)) \big| ,
\end{equation}
where $\Omega$ denotes all pixels within the instance mask, $I_t$ denotes the image at frame $t$, and $F_{t\rightarrow t+1}$ is the optical flow from frame $t$ to $t{+}1$.

\paragraph{Evaluating Spatial Consistency.}
We evaluate spatial rigidity using the \emph{divergence of the velocity field}, which measures local expansion and compression within the foreground region:
\begin{equation}
E_{\text{div}} = \frac{1}{|\Omega|}\sum_{p\in\Omega} \big| \nabla_x u(p) + \nabla_y v(p) \big| ,
\end{equation}
where $(u,v)=F_{t\rightarrow t+1}(p)$ are the flow components. Lower divergence indicates stronger preservation of rigid-body geometry. All computations are performed within instance masks \cite{ravi2024sam2segmentimages} using modern optical-flow estimators~\cite{wang2024sea}.

\subsubsection{Momentum and Energy Conservation}
The content generated by world models should preserve {momentum} and {energy} within the collision window; otherwise the impact fails to respect conservation laws. We therefore apply three checks: (1) \emph{linear momentum conservation}, (2) \emph{angular momentum conservation} and (3) \emph{kinetic energy conservation}. Together, these indicators assess whether motion transitions around impact remain consistent with vehicle dynamics and collision physics, verifying that generated behaviors follow physically plausible laws under dominant inter-vehicle impulses and negligible external forces.

\paragraph{(1) Linear Momentum Conservation.}  
For each vehicle \(i\) with mass \(m_i\) and pre-/post-impact velocities \(v_i^{-}\) and \(v_i^{+}\), the total momentum before and after collision is \(p^{-}=\sum_i m_i v_i^{-}\) and \(p^{+}=\sum_i m_i v_i^{+}\). Their difference \(\Delta p=p^{+}-p^{-}\) measures translational imbalance, and the normalized residual  
\begin{equation}
J_p=\frac{\|\Delta p\|}{\sum_i m_i\|v_i^{-}\|}
\end{equation}
quantifies momentum loss or gain relative to pre-impact motion. The evaluation is performed within a short collision window \(\Delta t\), during which external forces (e.g., tire friction or aerodynamic drag) contribute negligible impulse compared with inter-vehicle contact. Under this quasi-closed condition, linear momentum can be treated as approximately conserved \cite{stronge2018impact}. Small \(J_p\) values indicate physically consistent velocity transitions dominated by a single collision impulse, while larger values imply missing or unbalanced dynamics. 

\paragraph{(2) Angular Momentum Conservation.}  
We evaluate rotational consistency under the same impulse-window assumption by computing angular momentum about the contact point \(c\). For each vehicle \(i\) with yaw moment of inertia \(I_{z,i}\), yaw rate \(\omega_i\), position \(r_i\), and velocity \(v_i\), the total angular momentum is  
\begin{subequations}
\begin{align}
H_c &= \sum_{i=1}^{n} \!\left(I_{z,i}\omega_i 
      + m_i (r_i - c)\!\times\! v_i\right), \label{eq:Hc}\\[3pt]
J_H &= \frac{\|H_c^{+}-H_c^{-}\|}
            {\|H_c^{-}\|+\varepsilon}. \label{eq:JH}
\end{align}
\end{subequations}
Here \((r_i-c)\!\times\! v_i\) denotes the scalar 2D cross product capturing the rotational effect of translational motion around \(c\). Low \(J_H\) values indicate that angular momentum transfer is physically consistent with rigid-body impact dynamics, while large values imply unbalanced or nonphysical torque generation. The small constant $\varepsilon$ prevents division by zero when pre-impact angular momentum is negligible, ensuring numerical stability. 

\paragraph{(3) Kinetic Energy Conservation.}  
We verify that total kinetic energy does not increase during impact, consistent with the dissipative nature of real collisions. The total kinetic energy and its normalized residual are
\begin{subequations}
\begin{align}
E_k &= \tfrac{1}{2}\sum_{i=1}^{n} m_i \|v_i\|^2, \label{eq:Ek}\\[3pt]
J_E &= \max\!\left(0,\, \frac{E_k^{+}-E_k^{-}}{E_k^{-}}\right). \label{eq:JE}
\end{align}
\end{subequations}
Here $J_E$ serves as a penalty that activates only when kinetic energy increases; $J_E=0$ for $E_k^{+}\le E_k^{-}$ and grows proportionally with any unphysical energy gain.

\subsubsection{World-Dynamics Integrity}
We evaluate {world-dynamics integrity} to test whether each instance maintains a stable identity and consistent appearance through impact, ensuring coherent behavior at the instance level. Intuitively, 
a physically plausible collision should preserve the integrity of object identities and avoid sudden visual changes unexplained by the generated dynamics.
We examine two complementary dimensions: \emph{Instance Stability}, which measures the temporal persistence of tracked instances, and \emph{Appearance Drift}, which evaluates the smoothness and temporal coherence of instance-level visual features over the trajectory.

\paragraph{Instance Stability.}
We evaluate instance stability by measuring how consistently an instance retains the same identity throughout the collision. We implement this using the Simpson index: let $c_{k,i}$ denote the number of frames in which instance $i$ is assigned ID $k$, $N_i=\sum_k c_{k,i}$, and $f_{k,i}=c_{k,i}/N_i$. The score is
\begin{equation}
S_{\mathrm{ID}}^{(i)}=\sum_k f_{k,i}^2 \in [0,1],
\end{equation}
where $1$ indicates a fully stable identity and lower values reflect fragmentation. Tracking associations are obtained using CenterTrack \cite{zhou2020tracking}.

\paragraph{Appearance Drift.}
To capture appearance over time, we use SAM2 to obtain instance masks \cite{ravi2024sam2segmentimages}, and then assess instance-level visual coherence by tracking how the CLIP features of each vehicle evolve along its full visible trajectory.
For each instance $i$, let $\{X_{i,t}\}$ denote the sequence of masked crops and let $\hat z_{i,t}=\phi(X_{i,t})/\|\phi(X_{i,t})\|_2$ be the L2-normalized CLIP embedding at frame $t$.  
Given the ordered embedding path $\{\hat z_{i,t_1},\hat z_{i,t_2},\dots,\hat z_{i,t_T}\}$, we compute the frame-to-frame angular deviation using $\theta_{i,k}=\arccos(\langle \hat z_{i,t_k},\,\hat z_{i,t_{k+1}}\rangle)$ for consecutive frames $k$.  
The appearance drift score for instance $i$ is defined as
\begin{equation}
D_{\mathrm{ad}}^{(i)}
= \frac{1}{T-1}\sum_{k=1}^{T-1}\theta_{i,k},
\end{equation}
with lower values indicating smoother and more temporally stable appearance evolution.
The video-level $D_{\mathrm{ad}}$ is computed as a frame-count-weighted average over all tracked instances.

\begin{figure*}[t]
    \centering
        \includegraphics[width=0.95\linewidth]{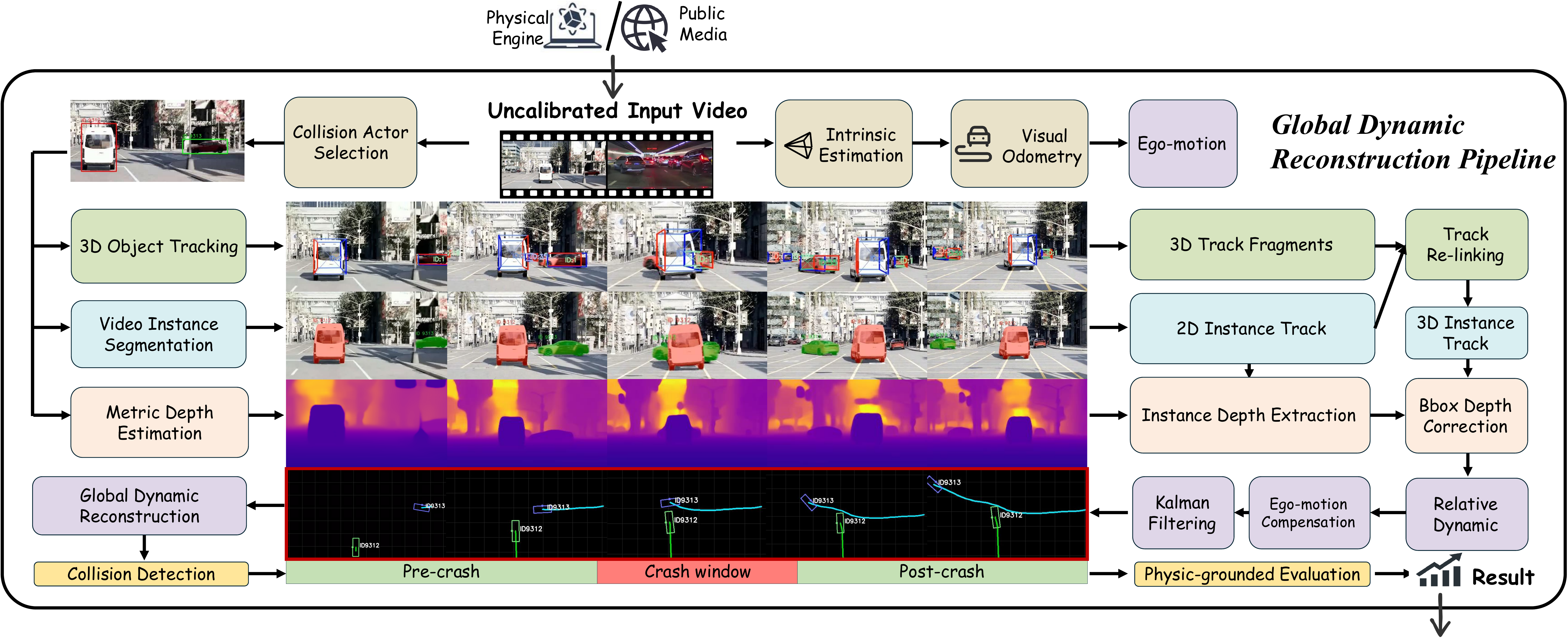}
    \caption{\textbf{Global dynamic reconstruction pipeline.} The system reconstructs the 3D dynamics of collision participants from uncalibrated accident videos, where camera parameters, scene scale, and ego-motion are not directly available. We combine 3D tracking, video segmentation, metric depth estimation, and visual odometry to obtain coherent per-actor trajectories. Tracking fragments are re-linked to maintain identity continuity, metric depth restores real-world scale, and ego-motion compensation maps relative motion into a global coordinate system. The resulting global trajectories define the pre-crash, crash, and post-crash window and support reliable computation of physical quantities used in our physics-based evaluation.}
    \label{fig:eval_pipeline}
    \vspace{-4mm}
\end{figure*}

\subsection{\BenchmarkName~Data Collection and Composition}
\label{sec:data}
To evaluate collision understanding of world models under both controllable synthetic settings and in-the-wild conditions, we construct CrashTwin by combining a CARLA-based crash generator with a curated real-world crash corpus.

\paragraph{Collision Data Collection Pipeline.}
Our collision data collection pipeline includes both simulated and real-world sources. In simulation, we reproduce seven high-frequency intersection collision types based on national pre-crash statistics~\cite{najm2007precrash}. Using CARLA, built on Unreal Engine~5, \BenchmarkName~simulates scalable, high-fidelity scenes where vehicles approach from diverse directions and speeds. Each case is parameterized by initial position, velocity, and orientation, with stochastic perturbations that vary timing and contact geometry. The simulator enforces rigid-body dynamics with frictional contact and records full logs of vehicle states, 3D bounding boxes, segmentation, depth, and optical flow, enabling reproducible, physics-based evaluation.
For real-world data, we collect diverse crash videos from public online sources and clean the raw data to obtain crash videos in varied environments such as intersections, highways, and rural roads. Each video is temporally segmented around the impact moment and annotated with detailed captions describing vehicle motion and scene context. Together, these simulated and real-world streams form a comprehensive and physically interpretable foundation for benchmarking collision understanding.

\begin{table}[t]
\centering
\scriptsize
\renewcommand{\arraystretch}{1.08}
\setlength{\tabcolsep}{2.0pt}

\caption{\textbf{Accident video dataset comparison.} We summarize prior datasets by release year, scale (\#clips and \#frames), whether object bounding boxes are provided (Bboxes), whether tracklets are provided (Tracklet), whether text descriptions are provided (TT), and whether the data are real or synthetic (R/S).}

\begin{tabular*}{0.96\columnwidth}{@{\extracolsep{\fill}}lccccccc@{}}
\toprule
Dataset & Year & \#Clips & \#Frames & Bboxes & Tracklet & TT & R/S \\
\midrule
DAD \cite{DBLP:conf/accv/ChanCXS16} & 2016 & 1,750  & 175K  &        & \checkmark &        & R \\
A3D \cite{DBLP:conf/iros/YaoXWCA19} & 2019 & 3,757  & 208K  &        &        & \checkmark & R \\
GTACrash \cite{DBLP:conf/aaai/KimLHS19} & 2019 & 7,720  & --    &        &        & \checkmark & S \\
VIENA$^2$ \cite{aliakbarian2019viena} & 2019 & 15,000 & 2.25M &        &        & \checkmark & S \\
CTA \cite{DBLP:conf/eccv/YouH20} & 2020 & 1,935  & --    &        &        & \checkmark & R \\
CCD \cite{bao2020uncertainty} & 2021 & 1,381  & 75K   &        & \checkmark & \checkmark & R \\
TRA \cite{DBLP:journals/tits/LiuLCLX22} & 2022 & 560    & --    &        &        & \checkmark & R \\
DADA-2000 \cite{DBLP:journals/tits/FangYQXY22} & 2022 & 2,000  & 658K  &        &        & \checkmark & R \\
DoTA \cite{yao2022dota} & 2022 & 5,586  & 732K  & partial &        & \checkmark & R \\
ROL \cite{karim2022attention} & 2023 & 1,000  & 100K  &        &        & \checkmark & R \\
DeepAccident \cite{DBLP:journals/corr/abs-2304-01168} & 2023 & --     & 57K   &        &        & \checkmark & S \\
CTAD \cite{luoICASSP2023} & 2023 & 1,100  & --    &        &        & \checkmark & S \\
MM-AU \cite{Fang_2024_CVPR} & 2023 & 11,217 & 2.19M & \checkmark &        & \checkmark & R \\
Nexar \cite{moura2025nexardashcamcollisionprediction} & 2025 & 1,500 & 1.70M &        &        & \checkmark & R \\
\textbf{Ours} & \textbf{2026} & \textbf{38,349} & \textbf{7.04M} & \checkmark & \checkmark & \checkmark & \textbf{Both} \\
\bottomrule
\end{tabular*}

{\scriptsize \textbf{Bboxes} object bounding boxes. \textbf{Tracklet} short-term bounding-box trajectories with identity linkage. \textbf{TT} text descriptions. \textbf{R/S} real or synthetic.}

\label{tab:accident_compare}
\end{table}

\begin{table*}[t]
\centering
\scriptsize
\setlength{\tabcolsep}{3pt}
\renewcommand{\arraystretch}{1.10}
\caption{\textbf{\BenchmarkName~Evaluation Leaderboard.} Open-source models are evaluated on CrashTwin-Eval, while proprietary models are evaluated on CrashTwin-Eval-Mini due to API and rate-limit constraints. Metrics are grouped into three physical families. Lower is better unless marked with $\uparrow$.}

\begin{tabular}{l|cc|ccc|cc}
& \multicolumn{2}{c|}{\colorbox{orange!10}{\strut\makecell{Spatio-temporal\\Consistency}}} &
\multicolumn{3}{c|}{\colorbox{yellow!10}{\strut\makecell{Momentum \& Energy\\Conservation}}} &
\multicolumn{2}{c}{\colorbox{green!10}{\strut\makecell{World-dynamics\\Integrity}}} \\
\cline{1-8}
\textbf{Models} &
{\fontsize{6.8pt}{7.2pt}\selectfont \makecell{\rule{0pt}{2.7ex}Warp\\Error}} &
{\fontsize{6.8pt}{7.2pt}\selectfont \makecell{\rule{0pt}{2.7ex}Flow\\Divergence}} &
{\fontsize{6.8pt}{7.2pt}\selectfont \makecell{\rule{0pt}{2.7ex}Momentum\\Residual}} &
{\fontsize{6.8pt}{7.2pt}\selectfont \makecell{\rule{0pt}{2.7ex}Angular\\Residual}} &
{\fontsize{6.8pt}{7.2pt}\selectfont \makecell{\rule{0pt}{2.7ex}Energy\\Gain}} &
{\fontsize{6.8pt}{7.2pt}\selectfont \makecell{\rule{0pt}{2.7ex}Instance\\Stability}} &
{\fontsize{6.8pt}{7.2pt}\selectfont \makecell{\rule{0pt}{2.7ex}App.\\Drift}} \\
\cline{2-8}
& {\fontsize{6.8pt}{7.2pt}\selectfont $E_{\mathrm{warp}} \downarrow$} &
  {\fontsize{6.8pt}{7.2pt}\selectfont $E_{\mathrm{div}} \downarrow$} &
  {\fontsize{6.8pt}{7.2pt}\selectfont $J_p \downarrow$} &
  {\fontsize{6.8pt}{7.2pt}\selectfont $J_H \downarrow$} &
  {\fontsize{6.8pt}{7.2pt}\selectfont $J_E \downarrow$} &
  {\fontsize{6.8pt}{7.2pt}\selectfont $S_{\mathrm{ID}} \uparrow$} &
  {\fontsize{6.8pt}{7.2pt}\selectfont $D_{\mathrm{ad}} \downarrow$} \\
\hline

\rowcolor{blue!5}
\multicolumn{8}{l}{\textit{Open-Source Models}} \\
SkyReel-1.3B~\cite{chen2025skyreelsv2infinitelengthfilmgenerative} & 0.0227 & 0.6103 & 0.9566 & 0.9628 & 0.9457 & 0.6660 & 0.3592 \\
Wan 2.1-14B~\cite{wan2025} & 0.0179 & 0.6320 & 0.8235 & 0.8494 & 0.7864 & 0.6760 & 0.3117 \\
Wan 2.2-5B~\cite{wan2025} & 0.0145 & \best{0.5959} & 0.8899 & 0.8975 & 0.8649 & \best{0.7254} & \best{0.3109} \\
Cosmos-Predict2-2B~\cite{nvidia2025cosmosworldfoundationmodel} & 0.0240 & 0.6748 & 0.8890 & 0.8954 & 0.8590 & 0.6129 & 0.3462 \\
Cosmos-Predict2-14B~\cite{nvidia2025cosmosworldfoundationmodel} & \best{0.0117} & 0.7180 & \best{0.6828} & \best{0.7629} & \best{0.6047} & 0.6737 & 0.3327 \\
\hline

\rowcolor{blue!5}
\multicolumn{8}{l}{\textit{Proprietary Models}} \\
Google Veo 3.1~\cite{google2025veo31} & \best{0.0097} & 0.6202 & 0.7743 & 0.8011 & 0.7460 & \best{0.7232} & 0.3075 \\
Hailuo 2.3~\cite{minimax2025hailuo} & 0.0151 & 0.6143 & \best{0.7664} & \best{0.7812} & 0.7285 & 0.6203 & \best{0.2968} \\
Seedance V1 Pro~\cite{gao2025seedance} & 0.0166 & \best{0.6130} & 0.7725 & 0.7837 & \best{0.7209} & 0.6007 & 0.3002 \\
\hline
\end{tabular}

\label{tab:crashtwin_leaderboard}
\end{table*}

\paragraph{Data Composition and Statistics.}
Our final collection includes approximately 25K synthetic and 12K real-world crash sequences. The synthetic subset covers 7 representative intersection collision configurations derived from national pre-crash typologies, spanning diverse approach angles, impact geometries, and environmental conditions such as weather and illumination. Each synthetic event is recorded at 30~Hz with full state logs, segmentation, depth, and optical flow, ensuring frame-level physical observability.
The real-world subset contains 12K curated crash clips sourced from online videos, featuring varied environments including urban intersections, highways, and rural roads. Each clip is temporally segmented around the impact frame and captioned to describe vehicle motion, collision timing, and scene context. Together, these two complementary sources provide a balanced corpus that combines physically controlled synthetic data for quantitative evaluation with visually rich real-world data for assessing realism and generalization. We provide further details and a few data samples in the \underline{supplementary materials}. A detailed comparison with prior accident video datasets is summarized in~\cref{tab:accident_compare}.

\subsection{Global Dynamic Reconstruction Pipeline.}
\label{sec:eval}
To consistently evaluate both generated crash content and web-scale real videos, which are typically uncalibrated, we build a calibration-free pipeline that reconstructs the 3D dynamics of the collision participants from raw video, as illustrated in~\cref{fig:eval_pipeline}. Conventional 2D/3D detection and tracking pipelines often fail under the large viewpoint changes, motion blur, and occlusions present in crash footage, making them unreliable for physics analysis. Our pipeline instead integrates multiple vision foundation models which remains robust under rapid motion and provides metric trajectories for subsequent physics-based evaluation.

\paragraph{Actor Selection.}
For each collision, we must first determine the pair of vehicles that will collide, using the initial frame of the sequence. Automatic identification of collision participants is not yet reliable for current vision-language models, especially under occlusion and rapid motion. We therefore use simulator ground-truth identities in synthetic clips, and annotate collision actors manually in real crash videos. These identities are released with the test set to ensure consistent actor selection across all evaluation methods.

\paragraph{Tracking Fragments.}
We obtain initial 3D track fragments for the selected actors using a 3D object tracker. These fragments provide coarse estimates of the underlying dynamics but often break under strong acceleration, occlusion, and rapid directional changes that are typical of collision events.

\paragraph{Relinking and Correction.}
To obtain coherent trajectories, fragmented 3D segments are relinked using video instance segmentation, which provides more reliable identity continuity under rapid motion. Metric depth is then applied to refine geometry and restore real-world scale, yielding stable relative 3D dynamics in the camera coordinate frame.

\paragraph{Global Dynamics.}
Relative trajectories are transformed into a global coordinate system by composing them with ego motion estimated through visual odometry. A Kalman filter is then applied to smooth positions and orientations, producing stable global dynamics for both actors.

\paragraph{Collision Detection.}
Given the reliable global trajectories, we monitor the distance between the two actors and mark the first moment it falls within a safety threshold, which defines the start of the collision window. A short pre-crash, crash, and post-crash segment is then extracted and passed to our physics-based evaluation. See more details in our supplementary materials.

\paragraph{Benchmark Protocol.}
For all evaluations, we estimate camera intrinsics with MapAnything~\cite{keetha2025mapanything}, obtain ego motion using DROID-SLAM~\cite{teed2022droidslamdeepvisualslam} combined with metric-scale correction from Metric3D V2~\cite{hu2024metric3d}, generate initial 3D object tracks with CenterTrack~\cite{zhou2020tracking}, and use SAM2~\cite{ravi2024sam2segmentimages} to produce instance masks for relinking and refinement. These components together provide the 3D trajectories required by the physics-based metrics in Section~\ref{sec:phys}. Collision timing is defined as the first frame in which the inter-vehicle distance falls below a predefined safety threshold. For clips where no contact occurs, collision-based momentum and energy metrics can become numerically unstable, so we assign a fixed penalty value in all non-collision cases.

\begin{figure}[t]
    \centering
        \includegraphics[width=0.95\linewidth]{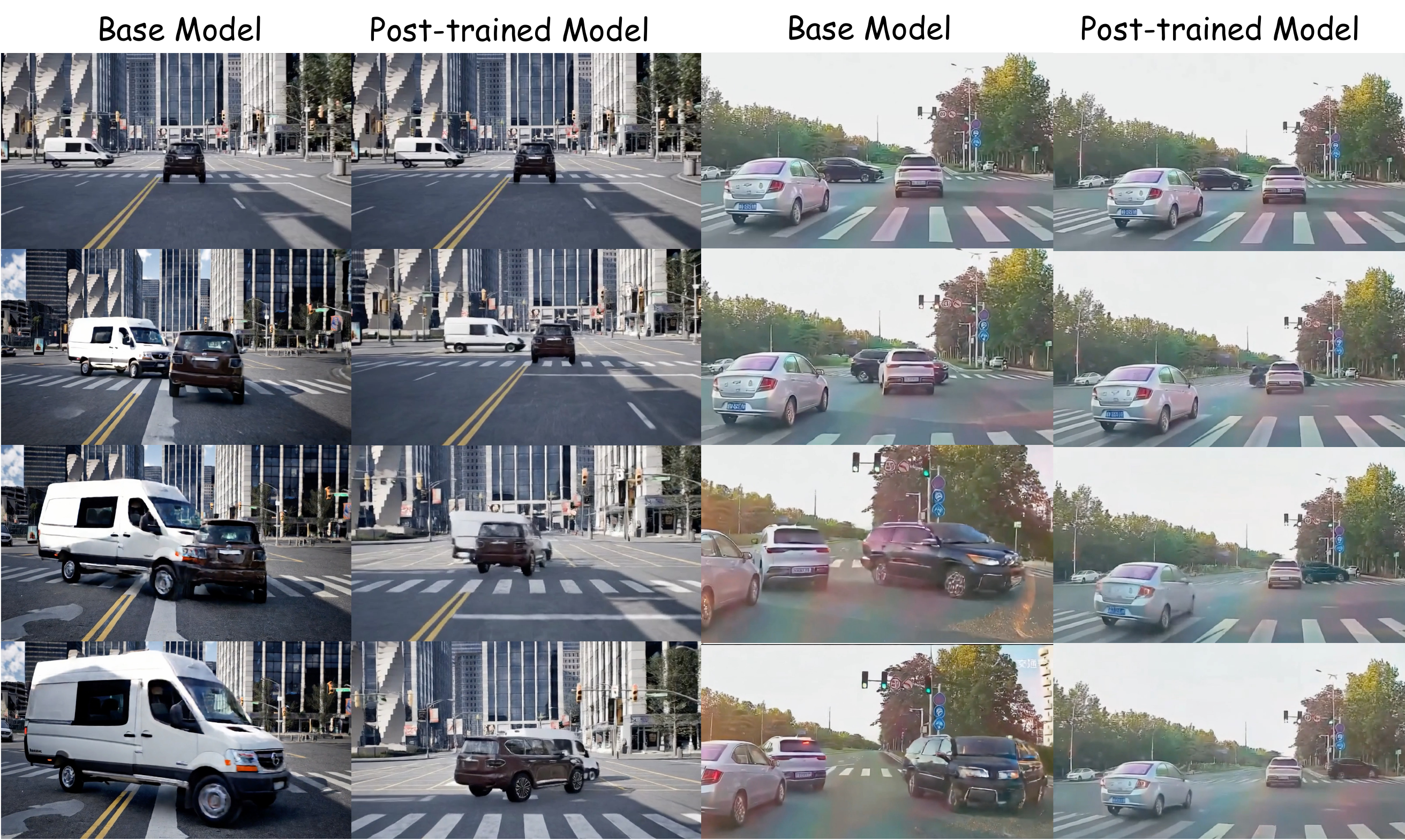}
    \caption{\textbf{Qualitative comparison before and after physics post-training.} 
\textit{Left:} The base model lets the van pass through the other vehicle without transferring lateral momentum, while the post-trained model produces a realistic rightward push. 
\textit{Right:} In a mild real crash, the base model shows appearance distortion and deformation, whereas the post-trained model preserves stable geometry and identity.}

    \label{fig:posttrain}
\end{figure}

\section{Experiments}
\subsection{Datasets}
To study how current world models perform on physically realistic collisions and how to reduce the gap between generated and real crash videos, we benchmark them on the proposed \BenchmarkName~test split under our evaluation protocol. As summarized in Tab.~\ref{tab:accident_compare}, prior accident video datasets are restricted in scale and confined to a single data source, lacking both the volume and cross-domain diversity necessary for unified evaluation and model adaptation. To bridge this gap, \BenchmarkName~contains 38K crash events, including both 25.6K synthetic and 12.6K real-world sequences, and is divided into a training split for model development and a held-out test split for evaluation.

We construct a comprehensive evaluation set, \textit{\BenchmarkName-Eval}, comprising 300 synthetic and 44 real-world videos. Each clip provides two annotated collision actors in the first frame and a textual description of the impact configuration. To accommodate closed-source models with high inference cost or limited access quotas, we further define a smaller evaluation partition, \textit{\BenchmarkName-Eval-Mini}, which contains randomly sampled 100 synthetic and 16 real-world videos. Both subsets are sampled from the \BenchmarkName~test split and are used for quantitative benchmarking, while the remaining data serve as training samples. We report the physics-grounded metrics defined in Section~\ref{sec:phys}.

\begin{table}[t]
\centering
\scriptsize
\setlength{\tabcolsep}{4pt}
\renewcommand{\arraystretch}{1.10}
\caption{\textbf{Reconstruction accuracy on simulated crashes.} Since no publicly accessible models exist for uncalibrated collision scenarios, we incrementally add components to a basic tracking baseline. We report mean absolute trajectory error (ATE) in 3D for both scene-level and instance-level dynamics.}
\label{tab:reconstruction_accuracy}
\begin{tabular*}{\columnwidth}{@{\extracolsep{\fill}}lcccc@{}}
\toprule
\multirow{2}{*}{\textbf{Model Configuration}} &
\multicolumn{2}{c}{\textbf{Scene-level ATE (m) $\downarrow$}} &
\multicolumn{2}{c}{\textbf{Instance-level ATE (m) $\downarrow$}} \\
\cmidrule(lr){2-3}\cmidrule(lr){4-5}
& \textbf{$SE(3)$} & \textbf{$Sim(3)$} & \textbf{$SE(3)$} & \textbf{$Sim(3)$} \\
\midrule
Basic 3D Tracking~\cite{zhou2020tracking} & 11.71 & 9.38 & 3.89 & 1.98 \\
+ Instance Relinking & 5.96 & 5.10 & 3.73 & 1.89 \\
+ Kalman Filtering & 5.61 & 4.68 & 3.29 & 1.47 \\
+ Metric Depth Correction & \textbf{5.48} & \textbf{3.70} & \textbf{2.63} & \textbf{0.91} \\
\bottomrule
\end{tabular*}
\end{table}

\subsection{Leaderboard Comparison.}
We benchmark representative world models, including the open-source SkyReel-1.3B~\cite{chen2025skyreelsv2infinitelengthfilmgenerative}, Wan 2.1~\cite{wan2025}, and Cosmos-Predict2~\cite{nvidia2025cosmosworldfoundationmodel}, on the \BenchmarkName-Eval subset, as well as proprietary models Veo 3.1~\cite{google2025veo31}, Hailuo~\cite{minimax2025hailuo}, and Seeddance~\cite{gao2025seedance}, on the \BenchmarkName-Eval-Mini subset.
~\cref{tab:crashtwin_leaderboard} shows that momentum and energy residuals remain noticeably high across all methods, indicating that current models still struggle to exchange momentum and energy in a physically consistent way, even when some achieve relatively strong spatio-temporal consistency and world-dynamics integrity.

As illustrated by the Hailuo 2.3 example in~\cref{fig:failcase}, the scene remains temporally stable with consistent identities, yet the blue truck suddenly aligns with the sedan after impact, causing its lateral momentum to vanish.
In the Seedance V1 Pro case, the police vehicle preserves appearance and identity throughout the sequence, but undergoes an implausible yaw reversal after contact, revealing a clear angular-momentum inconsistency. The Veo 3.1 example further shows two vehicles nearly coming to rest at impact before accelerating again, producing an unphysical increase in kinetic energy despite otherwise coherent frames. This gap indicates that maintaining spatio-temporal smoothness and stable object identities is easier for current world models than producing physically correct momentum and energy exchanges. It highlights the necessity of physics-grounded evaluation to reveal failures that perceptual metrics cannot capture.

\begin{table}[t]
\centering
\scriptsize
\setlength{\tabcolsep}{3pt}
\renewcommand{\arraystretch}{1.10}
\caption{\textbf{Effect of post training.} Metrics are grouped into three physical families. Lower is better unless marked with $\uparrow$.}
\label{tab:posttrain}

\begin{tabular}{@{\extracolsep{\fill}}l|cc|ccc|cc@{}}
& \multicolumn{2}{c|}{\colorbox{orange!10}{\strut\makecell{Spatio-temporal\\Consistency}}} &
  \multicolumn{3}{c|}{\colorbox{yellow!10}{\strut\makecell{Momentum \& Energy\\Conservation}}} &
  \multicolumn{2}{c}{\colorbox{green!10}{\strut\makecell{World-dynamics\\Integrity}}} \\
\cline{1-8}
\textbf{Model Variant} &
{\fontsize{6.5pt}{7.0pt}\selectfont \makecell{\rule{0pt}{2.6ex}Warp\\Error}} &
{\fontsize{6.5pt}{7.0pt}\selectfont \makecell{\rule{0pt}{2.6ex}Flow\\Div.}} &
{\fontsize{6.5pt}{7.0pt}\selectfont \makecell{\rule{0pt}{2.6ex}Momentum\\Residual}} &
{\fontsize{6.5pt}{7.0pt}\selectfont \makecell{\rule{0pt}{2.6ex}Angular\\Residual}} &
{\fontsize{6.5pt}{7.0pt}\selectfont \makecell{\rule{0pt}{2.6ex}Energy\\Gain}} &
{\fontsize{6.5pt}{7.0pt}\selectfont \makecell{\rule{0pt}{2.6ex}Instance\\Stability}} &
{\fontsize{6.5pt}{7.0pt}\selectfont \makecell{\rule{0pt}{2.6ex}App.\\Drift}} \\
\cline{2-8}
& {\fontsize{6.5pt}{7.0pt}\selectfont $E_{\mathrm{warp}} \downarrow$} &
  {\fontsize{6.5pt}{7.0pt}\selectfont $E_{\mathrm{div}} \downarrow$} &
  {\fontsize{6.5pt}{7.0pt}\selectfont $J_p \downarrow$} &
  {\fontsize{6.5pt}{7.0pt}\selectfont $J_H \downarrow$} &
  {\fontsize{6.5pt}{7.0pt}\selectfont $J_E \downarrow$} &
  {\fontsize{6.5pt}{7.0pt}\selectfont $S_{\mathrm{ID}} \uparrow$} &
  {\fontsize{6.5pt}{7.0pt}\selectfont $D_{\mathrm{ad}} \downarrow$} \\
\hline
Base (w/o Post-training) & 0.0240 & 0.6748 & 0.8890 & 0.8954 & 0.8590 & 0.6129 & 0.3462 \\
Base (w/ Post-training)  & 0.0085 & 0.6279 & 0.6479 & 0.7296 & 0.5534 & 0.7746 & 0.2953 \\
\textit{Ground Truth Reference} & 0.0075 & 0.6549 & 0.3089 & 0.4620 & 0.2502 & 0.8010 & 0.2819 \\
\hline
\end{tabular}
\end{table}

\paragraph{Effect of Post-Training.}
We fine-tune pretrained models on our \BenchmarkName~training split using physical consistency objectives that regularize flow-based temporal coherence, spatial rigidity, and short-window momentum and energy residuals. Post-training brings clear gains across all physical dimensions. As seen in~\cref{tab:posttrain}, most spatio-temporal and world-dynamics metrics move closer to the ground-truth reference, and momentum consistency also improves.
This indicates that post-training improves spatio-temporal consistency and world-dynamics integrity more than momentum and energy conservation, reinforcing that spatio-temporal smoothness and stable identities are easier to achieve than physically correct momentum and energy exchange.
A representative qualitative example in~\cref{fig:posttrain} further shows that post-training restores lateral momentum transfer and stabilizes instance appearance.

\subsection{Effectiveness of the Proposed Reconstruction}

As no publicly accessible model can recover metric-scale dynamics from uncalibrated severe-collision videos, we build a basic 3D tracking baseline~\cite{zhou2020tracking} and incrementally add our components. ~\cref{tab:reconstruction_accuracy} shows large drift under heavy occlusion for the baseline, which is steadily reduced by instance relinking, Kalman filtering, and metric depth correction. We report $SE(3)$ and $Sim(3)$ errors to reflect accuracy with fixed scale and up to a global similarity transform. The final instance-level error reaches a sub-meter regime, supporting that our pipeline recovers stable kinematics for physics-grounded evaluation. Additional qualitative results are included in the \underline{supplementary materials}.

\section{Conclusions}
This paper has introduced a physics-based benchmark for world models in crash scenarios, enabling a fine-grained examination of how generative rollouts behave under real physical constraints.
We have demonstrated that our benchmark reveals systematic weaknesses in current models, particularly in momentum exchange, energy transfer, and post-impact dynamics, even when their visual predictions appear coherent.
This is achieved by decomposing crash dynamics into interpretable physical metrics and by providing a pipeline that highlights where and how physical violations arise around high-force events.
Our study shows that post-training reduces violations more in spatio-temporal consistency and world-dynamics integrity than in momentum and energy exchange, motivating stronger physics priors and more reliable dynamic modeling.
Altogether, \BenchmarkName~serves as a diagnostic tool that combines data curation, physical attribute estimation, and physics-grounded metrics to bring physical fidelity to the forefront of world-model evaluation, paving the way for future models that are not only visually plausible but physically trustworthy.

\clearpage
\newcounter{phaisuppsectionanchor}
\pretocmd{\section}{\stepcounter{phaisuppsectionanchor}}{}{}
\renewcommand{\theHsection}{supplement.\arabic{phaisuppsectionanchor}}
\renewcommand{\theHsubsection}{supplement.\arabic{phaisuppsectionanchor}.\arabic{subsection}}

\section{Technical Appendices and Supplementary Material}

\renewcommand\thesection{\Alph{section}} 

\setcounter{section}{0}

\begin{itemize}
    \item Section~\ref{sec:MetricDetails} details the proposed physics-based evaluation metrics and provides visualizations that show how violations of each metric manifest in generated rollouts and real crash videos
\item Section~\ref{sec:DataCreation} describes the curation of our dataset, including controllable synthetic collisions generated by a physically grounded engine and in-the-wild real-world traffic accidents collected from public media, along with their textual captions and the overall dataset composition and statistics.

    \item Section~\ref{sec:supp-global-recon} presents the global dynamic reconstruction pipeline for uncalibrated collision videos, covering 3D tracking fragments, metric depth refinement, ego-motion compensation, Kalman smoothing, and the extraction of pre-crash, crash, and post-crash windows.
\item Section~\ref{sec:supp-collision-types} analyzes performance by collision type, regrouping seven script-level intersection variants into three geometry-based collision families and reporting per-category physics-grounded metrics.

\item Section~\ref{sec:supp-robustness} reports additional robustness analyses, including evaluator calibration, reconstruction sensitivity, friction-aware and window-size ablations, synthetic-oracle reconstruction validation with runtime, and synthetic/real split evaluation.

\item Section~\ref{sec:humaneval} summarizes the 2AFC human study, including the protocol, basic statistics, and the annotation interface.

    \item Section \ref{sec:PostTrainingInference} outlines the post-training setup for adapting the video-to-world model to collision scenarios, summarizing the model configuration, training corpus, and distributed optimization strategy.
\end{itemize}

\section{Evaluation Metric Details}
\label{sec:MetricDetails}

\subsection{Physics-Grounded Metric Configuration}

\paragraph{Evaluation Protocol.}
All physics-grounded metrics are computed on reconstructed 3D trajectories and instance masks of the selected annotated collision actors in CrashTwin-Eval and CrashTwin-Eval-Mini. For open-source models, we evaluate on a fixed set of $344$ crash scenarios, including $300$ synthetic simulations and $44$ real-world videos. Proprietary models are assessed on a disjoint subset of $116$ scenarios ($100$ synthetic and $16$ real) due to API and rate-limit constraints. 
Before reconstruction, every rollout is converted to a common video format, resized to $1920\times1080$, and temporally resampled to $20$\,FPS. All metrics that involve temporal derivatives operate on this unified time grid in the common global coordinate frame produced by our pipeline.
The impact time is localized by detecting the first frame of the first valid contact event between the two reconstructed vehicles, where a frame $t$ is considered in contact when $\|p_1(t)-p_2(t)\|_2 \le 1.2(r_1+r_2)$, with $p_1(t)$ and $p_2(t)$ denoting the reconstructed 3D centers of the two vehicles and $r_i=\frac{1}{2}\|[h_i,w_i,l_i]\|_2$ denoting the effective radius of vehicle $i$ computed from the median height $h_i$, width $w_i$, and length $l_i$ of its 3D boxes along the trajectory, and this condition must hold for three consecutive frames; subsequent metrics are evaluated in short pre- and post-impact windows around this frame so that scores remain comparable across real and synthetic crashes, while samples with no valid contact event receive a fixed penalty of $1.0$ for each collision-dependent metric, namely $J_p$, $J_H$, and $J_E$.

\paragraph{Spatio-temporal consistency metrics.}
On the temporally normalized $20$\,FPS grid described above, we consider all consecutive frame pairs along the selected annotated vehicle tracks and estimate dense optical flow between the original RGB frames. All computations are restricted to per-frame instance masks predicted by SAM2~\cite{ravi2024sam2segmentimages} for the two actors, with a small erosion to reduce boundary artifacts, and frame pairs whose eroded masks contains fewer than 150 valid pixels are treated as invalid and skipped during metric computation.
 For $E_{\text{div}}$, we compute the divergence of the flow field and take the median absolute divergence inside the mask for each frame pair, then normalize the resulting per-pair series for each instance by its $95th$ percentile and clip it to $[0,1]$ before averaging over time and over the selected actors. The temporal warping error $E_{\text{warp}}$ reuses the same cached flow fields together with the temporal-warping routine and forward/backward consistency checks from~\cite{wang2024sea}, and we report the mean warping error over all valid frame pairs in each scenario.

\paragraph{Momentum and energy conservation metrics.}
For the momentum- and energy-based metrics, we adopt a fixed, symmetric collision window around the localized 3D impact frame. In all experiments we use 5 frames before and 5 frames after impact, which correspond to $0.25s$ on each side at $20$\,FPS, and we apply this configuration uniformly to all models and scenarios. Within this window, pre- and post-impact translational and rotational velocities are obtained from the Kalman-smoothed 3D trajectories and yaw angles.
Vehicle masses and inertial parameters for the two collision actors are fixed once the actors are selected, using simulator metadata for synthetic clips and category-level nominal estimates from VLM-identified vehicle types for real-world clips. The contact point $c$ is approximated at the first-contact frame from the two reconstructed actor boxes, and each vehicle's center of mass is taken as the box center. These quantities are then reused across all momentum-based metrics.
The residuals $J_p$ and $J_H$ are then computed, using these windowed pre- and post-impact quantities. The kinetic-energy term $J_E$ uses the same velocity estimates, normalizes the change in kinetic energy by the pre-impact energy with a small stabilizing constant, and clips the result to $[0,1]$ to avoid undue influence from rare large deviations.

\paragraph{World-dynamics integrity metrics.}
For the world-dynamics metrics, we keep the definitions of ID stability $S_{\text{ID}}$ and appearance drift $D_{\text{ad}}$ unchanged and specify only the implementation details. Tracker IDs are obtained by linking CenterTrack~\cite{zhou2020tracking} trajectories with SAM2~\cite{ravi2024sam2segmentimages} masks, and we include a dedicated label for missing detections so that gaps in tracking are treated consistently in the Simpson~\cite{simpson1949measurement} index used for $S_{\text{ID}}$. For $D_{\text{ad}}$, SAM2 masks are used to extract masked RGB crops for each actor; we discard frames whose mask covers less than $0.05\%$ of the image area and uniformly subsample at most $64$ valid frames per vehicle to limit redundancy. Each crop is resized to $224\times224$ and embedded using a CLIP image encoder ~\cite{Radford21ICML-CLIP} based on the ViT-B/32 architecture from OpenCLIP. We compute the angular distance between consecutive embeddings and average these angles along the trajectory to obtain a per-vehicle drift score, and the final $D_{\text{ad}}$ is a frame-count-weighted average of these scores over the selected collision actors.

\paragraph{Normalized metric display.}
To make small-valued metrics easier to compare across metric families, we provide a reference-anchored display of the main leaderboard in~\cref{tab:normalized_metric_display}. All columns are converted so that higher is better. The ground-truth reference defines a practical upper anchor of 90, and the weakest evaluated model defines 50. For $E_{\mathrm{div}}$, whose ground-truth reference is non-zero, we score closeness to this reference. This table is for display only, and all conclusions use the raw metric values.

\begin{table}[t]
\centering
\scriptsize
\setlength{\tabcolsep}{3pt}
\renewcommand{\arraystretch}{1.10}
\vspace{-4pt}
\caption{\textbf{Reference-anchored normalized display.}
Scores are normalized for readability, with higher values indicating better performance.}
\label{tab:normalized_metric_display}
\vspace{-2pt}
\begin{tabular}{l|cc|ccc|cc}
\multicolumn{1}{c|}{} &
\multicolumn{2}{c|}{\colorbox{orange!10}{\strut\makecell{Spatio-temporal\\Consistency}}} &
\multicolumn{3}{c|}{\colorbox{yellow!10}{\strut\makecell{Momentum \& Energy\\Conservation}}} &
\multicolumn{2}{c}{\colorbox{green!10}{\strut\makecell{World-dynamics\\Integrity}}} \\
\cline{1-8}
\textbf{Model} &
{\fontsize{6.8pt}{7.2pt}\selectfont \makecell{\rule{0pt}{2.7ex}Warp\\Error}} &
{\fontsize{6.8pt}{7.2pt}\selectfont \makecell{\rule{0pt}{2.7ex}Flow\\Divergence}} &
{\fontsize{6.8pt}{7.2pt}\selectfont \makecell{\rule{0pt}{2.7ex}Momentum\\Residual}} &
{\fontsize{6.8pt}{7.2pt}\selectfont \makecell{\rule{0pt}{2.7ex}Angular\\Residual}} &
{\fontsize{6.8pt}{7.2pt}\selectfont \makecell{\rule{0pt}{2.7ex}Energy\\Gain}} &
{\fontsize{6.8pt}{7.2pt}\selectfont \makecell{\rule{0pt}{2.7ex}Instance\\Stability}} &
{\fontsize{6.8pt}{7.2pt}\selectfont \makecell{\rule{0pt}{2.7ex}App.\\Drift}} \\
\cline{2-8}
& {\fontsize{6.8pt}{7.2pt}\selectfont $E_{\mathrm{warp}}$} &
{\fontsize{6.8pt}{7.2pt}\selectfont $E_{\mathrm{div}}$} &
{\fontsize{6.8pt}{7.2pt}\selectfont $J_p$} &
{\fontsize{6.8pt}{7.2pt}\selectfont $J_H$} &
{\fontsize{6.8pt}{7.2pt}\selectfont $J_E$} &
{\fontsize{6.8pt}{7.2pt}\selectfont $S_{\mathrm{ID}}$} &
{\fontsize{6.8pt}{7.2pt}\selectfont $D_{\mathrm{ad}}$} \\
\hline
SkyReel-1.3B & 53.2 & 61.7 & 50.0 & 50.0 & 50.0 & 63.0 & 50.0 \\
Wan 2.1-14B & 64.8 & 75.5 & 58.2 & 59.1 & 59.2 & 65.0 & 74.6 \\
Wan 2.2-5B & 73.0 & 52.6 & 54.1 & 55.2 & 54.6 & 74.9 & 75.0 \\
Cosmos-Predict2-2B & 50.0 & 77.4 & 54.2 & 55.4 & 55.0 & 52.4 & 56.7 \\
Cosmos-Predict2-14B & 79.8 & 50.0 & 66.9 & 66.0 & 69.6 & 64.6 & 63.7 \\
Google Veo 3.1 & 84.7 & 68.0 & 61.3 & 62.9 & 61.5 & 74.5 & 76.8 \\
Hailuo 2.3 & 71.6 & 64.3 & 61.7 & 64.5 & 62.5 & 53.9 & 82.3 \\
Seedance V1 Pro & 67.9 & 63.4 & 61.4 & 64.3 & 62.9 & 50.0 & 80.5 \\
\hline
\end{tabular}%
\vspace{-3pt}
\end{table}

The normalized view makes the cross-family tradeoffs more visible. 
Cosmos-Predict2-14B obtains the strongest momentum-and-energy scores, while Google Veo 3.1 and Hailuo 2.3 are more competitive on appearance-related metrics. 
This confirms that the benchmark exposes complementary failure modes rather than a single dominant visual-quality axis.

\begin{figure*}[t]
    \centering
     \includegraphics[width=\textwidth]{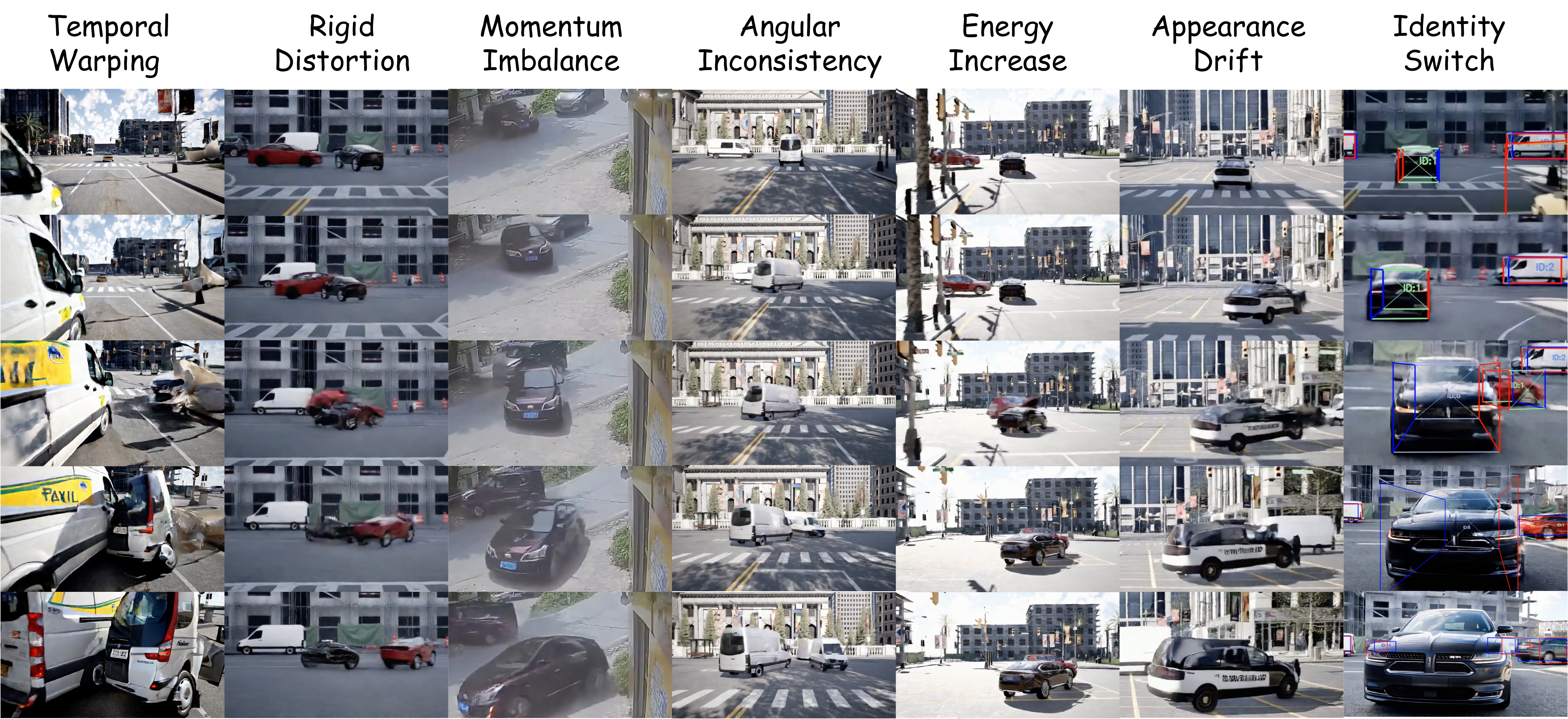}
    \caption{\textbf{Qualitative illustration of typical failure patterns captured by our physics-grounded evaluation metrics.} Each column visualizes a generated crash sequence that obtains a low score for the corresponding metric, covering temporal warping, rigid distortion, momentum imbalance, angular inconsistency, non-physical energy increase, appearance drift, and identity instability, and providing an intuitive reference for interpreting our physics-grounded evaluation.}
    \label{fig:metric_viz}
\end{figure*}

\paragraph{Qualitative intuition.}
To complement the formal definitions above and the aggregate scores reported in the main paper, we provide qualitative visualizations in~\cref{fig:metric_viz} that illustrate how typical low-score failures manifest in concrete crash rollouts. These examples justify the design of each physics-grounded metric, clarify which kinds of violations it is most sensitive to, and explain why the corresponding numerical values are meaningful diagnostic signals rather than abstract scalar quantities. We include these visual cases in the supplementary material because they require more space than is available in the main paper and provide additional interpretability for readers who wish to inspect the metrics more closely. We briefly summarize the characteristic error pattern shown in each column of~\cref{fig:metric_viz}:
\begin{itemize}[leftmargin=1.5em]
\item \textbf{Temporal warping.}
The first column of~\cref{fig:metric_viz} shows frames in which a severely distorted vehicle appears abruptly in free space, as if geometry from a future time step had been injected into an earlier frame while the rest of the scene remains in a pre-impact state. This violates temporal continuity and indicates a breakdown of frame-to-frame alignment.
\item \textbf{Rigid distortion.}
The second column illustrates non-physical soft-body deformation, where vehicles stretch, compress, or twist as if made of elastic material rather than rigid bodies. Both the red pickup and the white van exhibit near-complete geometric collapse under impact, which is inconsistent with realistic crash-induced damage.
\item \textbf{Momentum imbalance.}
In the third column, the striking vehicle retains nearly all of its pre-impact speed after contact, whereas the impacted vehicle exhibits very limited forward displacement or deflection. The resulting post-impact motion contradicts the expected redistribution of linear momentum between the two actors.
\item \textbf{Angular inconsistency.}
The fourth column shows a case where the contacted vehicle’s rear swings laterally while its front remains effectively fixed in place, so that the car appears to rotate around an incorrect pivot. This behavior is inconsistent with the torque direction and rotation center induced by the actual contact point.
\item \textbf{Energy increase.}
The fifth column depicts vehicles that exhibit exaggerated motion immediately after collision, accompanied by strong motion blur and enlarged displacements. This pattern reflects a non-physical increase in kinetic energy within the system instead of the expected dissipation.
\item \textbf{Appearance drift.}
The sixth column shows a vehicle whose rear geometry gradually changes from a sedan-like profile into a taller, SUV-shaped contour as the sequence progresses. Such temporal drift in shape, texture, and overall visual identity reveals long-horizon inconsistencies in appearance modeling for a single physical object.
\item \textbf{Identity instability.}
The last column visualizes identity switches, where the same black vehicle transitions from ID=1 to ID=6. These discontinuities indicate failures in tracking the same physical object with a consistent instance-level identity, motivating the design of the ID stability metric.
\end{itemize}
Together, these qualitative examples provide concrete reference points for interpreting the quantitative physics-grounded scores reported in later tables and figures, and they help relate metric values to recognizable physical and perceptual failure modes in generated crash videos.

\section{Data Curation}
\label{sec:DataCreation}

\paragraph{Synthetic collision videos.}
We instantiate seven intersection collision types derived from the NHTSA pre-crash typology using parametric CARLA~\cite{Dosovitskiy17CoRL-CARLA} scripts on multiple four-way and T-junction across the CARLA simulator’s Town 10 urban map. Each scenario contains two vehicles that undergo the collision and a third vehicle that simply follows behind as an observer. We enumerate combinations of vehicle models and turning or straight-line speeds between 25 and 50\,km/h in 5\,km/h increments, randomly shuffle all candidates with a fixed seed, and sub-sample about 34.3K scenarios; after simulation we discard cases where no collision occurs, yielding roughly 25K synthetic crash clips. These clips are included in the CrashTwin training split, and a 300-clip subset from the remainder forms the synthetic component of our core evaluation set, together with logged simulator states for the involved vehicles.

\paragraph{Real-world crash videos.}
In addition to the synthetic scenarios, we collect 12{,}683 real-world crash clips from public online platforms that host dashcam-like and traffic-camera footage. Since many uploads are compilation-style videos containing multiple accidents, we apply Gemini~2.5~Flash~\cite{comanici2025gemini} to automatically identify the start and end of each individual collision and extract a short clip around that event, yielding temporally localized crash snippets. We use 44 clips as the real component of our evaluation sets, while the remaining 12{,}639 clips are used in the CrashTwin training splits alongside the synthetic data. The real-world portion covers a broad range of road geometries and traffic conditions, including urban intersections, multilane highways, and lower-speed urban or rural segments.

\paragraph{Captions and metadata.}
Synthetic clips are annotated with text generated directly from simulator metadata to describe the collision type, involved vehicles, and physical parameters such as approach speeds. For real world clips, we apply Gemini~2.5~Flash to produce textual summaries that capture the scene context, participants, and pre impact motion. Beyond textual descriptions, the dataset metadata incorporates time varying kinematic action labels represented by per frame 3D bounding boxes for the colliding vehicles. These geometric trajectories implicitly encode the evolving physical state of the actors, capturing their precise position, orientation, velocity, and motion paths over time. While these physical signals are directly exported from the simulation engine for synthetic clips, they are recovered using our proposed pipeline for the real world clips. Together, the semantic captions and structured kinematic metadata provide a concrete foundation to support scenario understanding, rigorous evaluation, and detailed qualitative analysis.

\begin{figure*}[t]
    \centering
         \includegraphics[width=\textwidth]{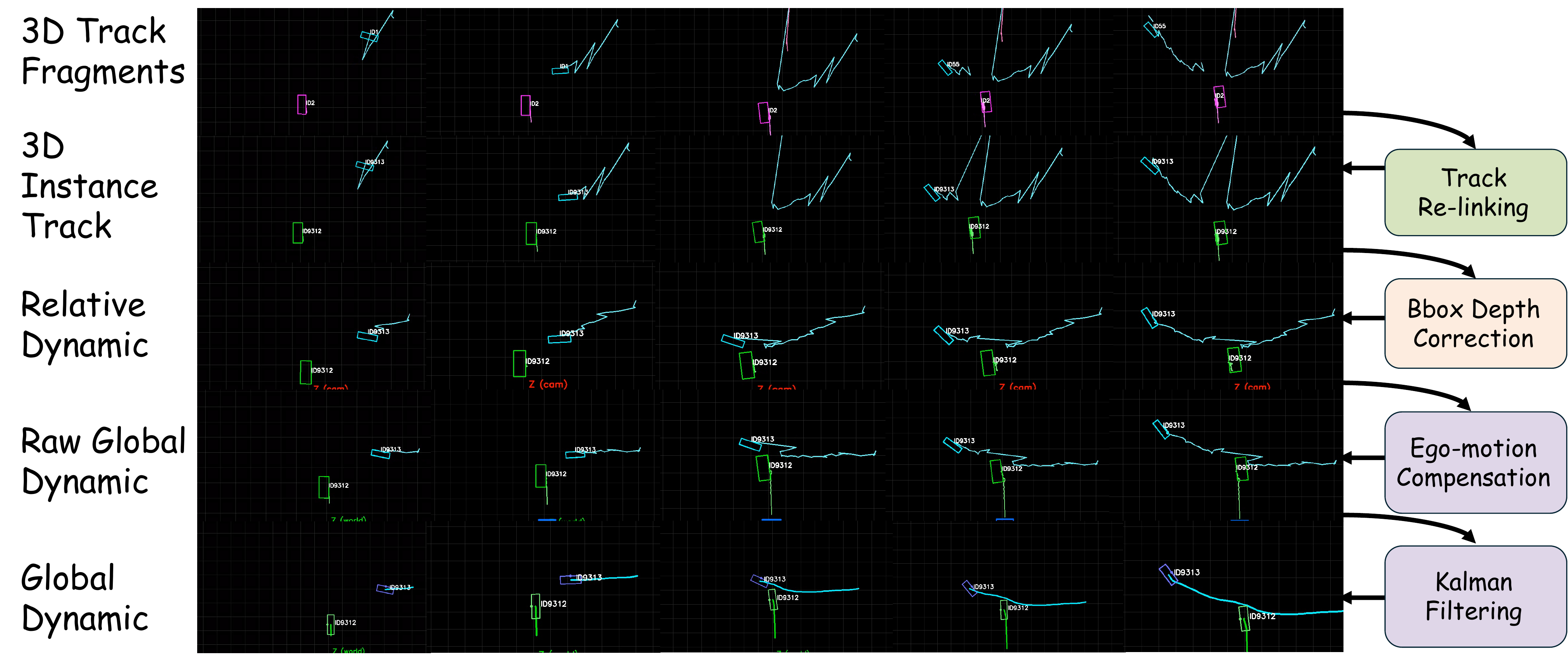}

\caption{\textbf{Global dynamic reconstruction stages.}
For a representative collision, we visualize the intermediate 3D trajectories after each stage of the reconstruction pipeline.
From top to bottom, the rows show initial 3D track fragments, re-linked 3D instance tracks guided by SAM2 masks, relative dynamics after metric depth correction, raw global trajectories after ego motion compensation, and the final Kalman-smoothed global dynamics in the world coordinate frame.
The figure illustrates how the pipeline progressively removes identity switches, depth-scale errors, and ego-motion artefacts to yield stable world-frame trajectories that are suitable for physics-based evaluation.}

    \label{fig:supp-relink-depth}
\end{figure*}

\section{Global Dynamic Reconstruction Methodology}
\label{sec:supp-global-recon}

This section describes the global dynamic reconstruction pipeline that underpins our physics-based metrics. Starting from an uncalibrated crash video and the two annotated collision actors, the pipeline recovers temporally smooth, metric 3D trajectories and yaw angles in a common world coordinate frame. Reconstruction proceeds in four stages: track re-linking to obtain temporally consistent 3D instance tracks, metric depth correction to recover physically meaningful relative motion, ego-motion compensation to map trajectories into a global frame, and Kalman filtering~\cite{kalman1960new} to suppress residual noise and outliers. The resulting trajectories serve as the sole input for the momentum, energy, and angular momentum scores in the main paper. ~\cref{fig:supp-relink-depth} visualizes these four stages and shows how the pipeline progressively converts fragmented, noisy tracks into robust global dynamics suitable for physics-based evaluation.

\subsection{Implementation Details}

We briefly summarize implementation choices that are not detailed in the main paper. In all cases we reconstruct only the two collision actors. For synthetic scenarios we use simulator-provided actor identifiers. For real crashes annotators select the two vehicles in the first frame; these identities are then fixed and used to filter both detector outputs and SAM2~\cite{ravi2024sam2segmentimages} mask tracks.

\paragraph{Track relinking.}
We use SAM2 video masks as long-range identity cues and treat monocular 3D detector tracks as candidates that can be merged. For each frame, we derive a tight 2D box from the SAM2 mask and compare it with the 2D projections of all detection boxes using an overlap score defined as the maximum of intersection-over-union and containment. Matches with overlap below a relaxed threshold (0.1) are ignored; the remaining matches are grouped into runs of consecutive frames, and we keep only runs whose per-frame overlap exceeds a stricter threshold (0.2) and whose length is at least six frames. We then greedily select the longest high-overlap runs and relabel their detections with a single tracking identifier tied to the SAM2 car, discarding short isolated matches as spurious. This procedure is applied only to the two annotated collision actors, so it resolves ID switches within their trajectories without changing which physical vehicles are evaluated, and corresponds to the second row of~\cref{fig:supp-relink-depth}, where the relinked instance tracks are visualized.

\paragraph{Metric depth correction.}
Depth maps are predicted with Metric3D~\cite{hu2024metric3d}. When a SAM2 mask is available, we randomly sample up to $K\!=\!2000$ pixels inside the mask; otherwise we sample inside the projected 2D detector box. Sampled depths are filtered by percentile clipping at the 5th and 95th percentiles and by an absolute range constraint $[1\,\mathrm{m}, 200\,\mathrm{m}]$. If fewer than 20 valid samples remain, the frame is marked unreliable and the original monocular depth is kept. Otherwise, we compute a truncated median of the remaining depths and update only the distance of the 3D box center along the viewing ray, leaving the 2D footprint, box size, and yaw unchanged. This improves metric scale and relative spacing between vehicles without introducing artificial changes in object shape, and corresponds to the third row of~\cref{fig:supp-relink-depth}, where the raw and depth-corrected trajectories are visualized.

\paragraph{Ego-motion composition and global coordinates.}
Camera motion is estimated with DROID-SLAM~\cite{teed2022droidslamdeepvisualslam}. Since SLAM is up to an unknown scale, we align it to metric depth by a least-squares fit between SLAM-inferred depths for static background points and Metric3D predictions. The first valid camera pose after SLAM initialization is fixed as the origin of a world coordinate system; the world $z$-axis is kept vertical. For each frame, the 3D box center in camera coordinates is transformed with the corresponding camera-to-world matrix to obtain raw global trajectories for both actors. We also transform finite-difference velocities, which are later used directly in momentum and energy computation.  This stage corresponds to the fourth row of~\cref{fig:supp-relink-depth}, where the raw global trajectories after ego-motion compensation are visualized.

\paragraph{Kalman smoothing of positions and yaw.}
Finally, we apply probabilistic smoothing to reduce frame-level jitter. For each actor we run a standard forward Kalman filter followed by a Rauch--Tung--Striebel (RTS) ~\cite{rauch1965maximum} backward pass on a constant-velocity state $\mathbf{x}_t = [\mathbf{p}_t, \mathbf{v}_t]^\top \in \mathbb{R}^6$ with white-noise acceleration. Process and observation covariances are chosen such that a displacement of roughly one vehicle length over five frames is treated as typical; the physics-based metrics are insensitive to moderate changes of these hyperparameters. Yaw angles are handled separately: we first unwrap them to a continuous sequence, run a one-dimensional Kalman filter on yaw and yaw rate, skip updates in frames where the 3D box yaw is unstable (strong occlusion or near-frontal views), and finally wrap the result back to $[-\pi,\pi]$. The smoothed world-frame positions and yaw sequences correspond to the bottom row of~\cref{fig:supp-relink-depth} and are the trajectories used in all momentum-, energy-, and angular-momentum-based scores.

\begin{figure*}[t]
    \centering
    \includegraphics[width=\textwidth]{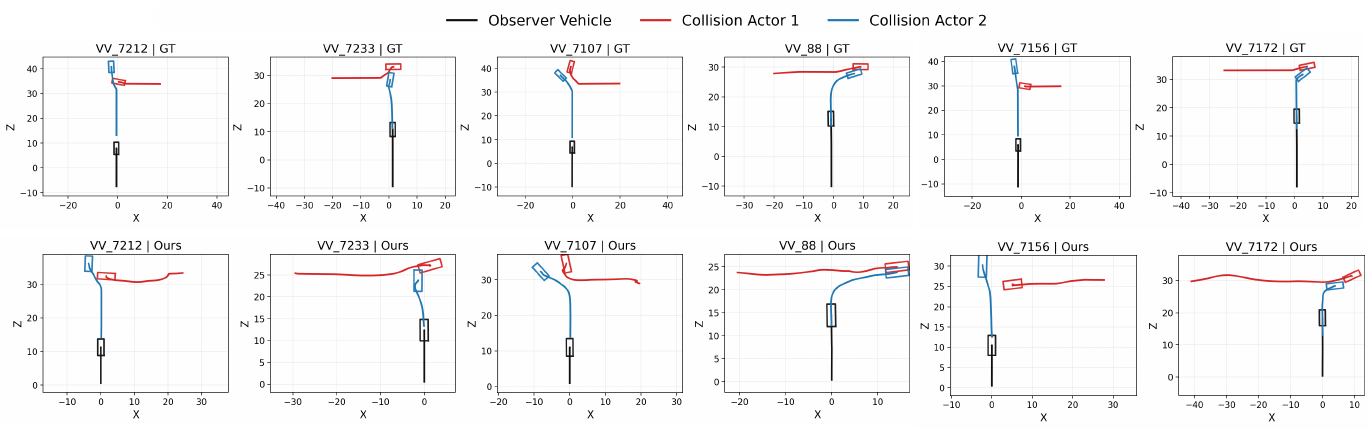}
    \caption{\textbf{Additional qualitative reconstruction results.}
    Each column corresponds to one scenario. The top row shows the ground-truth bird's-eye-view trajectories, and the bottom row shows our reconstructed trajectories. Black denotes the observer vehicle, while red and blue denote the two collision actors. Across diverse cases, the reconstructed trajectories preserve the overall interaction geometry, collision location, and coarse post-impact motion trends with good fidelity.}
    \label{fig:supp-recon-qual}
\end{figure*}

\subsection{Additional Qualitative Reconstruction Results}
\label{sec:supp-global-recon-qual}

To complement the stage-wise visualization in~\cref{fig:supp-relink-depth}, we provide additional qualitative reconstruction results on representative collision scenarios in~\cref{fig:supp-recon-qual}. Each column shows one scenario, with the top row denoting the ground-truth bird's-eye-view trajectories and the bottom row showing our reconstructed global trajectories. In all panels, the black trajectory denotes the observer vehicle, while the red and blue trajectories denote the two collision actors. These examples illustrate that the proposed reconstruction pipeline recovers the overall interaction geometry, including vehicle approach patterns, collision regions, and coarse post-impact motion trends, across diverse crash cases.

\begin{table}[t]
\centering
\scriptsize
\setlength{\tabcolsep}{2pt}
\renewcommand{\arraystretch}{1.10}
\vspace{-4pt}
\caption{\textbf{Per-collision-category physics-grounded metrics on synthetic intersections in CrashTwin-Eval.} 
For each collision family we compare a baseline world model and our method in terms of spatio-temporal consistency ($E_{\mathrm{warp}}\downarrow$, $E_{\mathrm{div}}\downarrow$), 
momentum and energy conservation ($J_p\downarrow$, $J_H\downarrow$, $J_E\downarrow$), 
and world-dynamics integrity ($S_{\mathrm{ID}}\uparrow$, $D_{\mathrm{ad}}\downarrow$). 
Lower values indicate better performance for all error terms and higher values are better for $S_{\mathrm{ID}}$. 
Right-angle crossing and opposing-path / head-on-like conflicts exhibit the largest gains in physics-grounded metrics, 
while shallow-angle sideswipe / glancing collisions remain the most challenging.}
\label{tab:collision-family-results}
\vspace{-2pt}

\begin{tabular}{@{\extracolsep{\fill}}p{2.8cm}l|cc|ccc|cc@{}}
& &
\multicolumn{2}{c|}{\colorbox{orange!10}{\strut\makecell{Spatio-temporal\\Consistency}}} &
\multicolumn{3}{c|}{\colorbox{yellow!10}{\strut\makecell{Momentum \& Energy\\Conservation}}} &
\multicolumn{2}{c}{\colorbox{green!10}{\strut\makecell{World-dynamics\\Integrity}}} \\
\cline{1-9}
\textbf{Collision category} &
\textbf{Model} &
{\fontsize{6.5pt}{7.0pt}\selectfont \makecell{\rule{0pt}{2.6ex}Warp\\Error}} &
{\fontsize{6.5pt}{7.0pt}\selectfont \makecell{\rule{0pt}{2.6ex}Flow\\Div.}} &
{\fontsize{6.5pt}{7.0pt}\selectfont \makecell{\rule{0pt}{2.6ex}Momentum\\Residual}} &
{\fontsize{6.5pt}{7.0pt}\selectfont \makecell{\rule{0pt}{2.6ex}Angular\\Residual}} &
{\fontsize{6.5pt}{7.0pt}\selectfont \makecell{\rule{0pt}{2.6ex}Energy\\Gain}} &
{\fontsize{6.5pt}{7.0pt}\selectfont \makecell{\rule{0pt}{2.6ex}Instance\\Stability}} &
{\fontsize{6.5pt}{7.0pt}\selectfont \makecell{\rule{0pt}{2.6ex}App.\\Drift}} \\
\cline{3-9}
& &
{\fontsize{6.5pt}{7.0pt}\selectfont $E_{\mathrm{warp}} \downarrow$} &
{\fontsize{6.5pt}{7.0pt}\selectfont $E_{\mathrm{div}} \downarrow$} &
{\fontsize{6.5pt}{7.0pt}\selectfont $J_p \downarrow$} &
{\fontsize{6.5pt}{7.0pt}\selectfont $J_H \downarrow$} &
{\fontsize{6.5pt}{7.0pt}\selectfont $J_E \downarrow$} &
{\fontsize{6.5pt}{7.0pt}\selectfont $S_{\mathrm{ID}} \uparrow$} &
{\fontsize{6.5pt}{7.0pt}\selectfont $D_{\mathrm{ad}} \downarrow$} \\
\hline
\multirow{2}{*}{\makecell[l]{Right-angle cross.}}
& Baseline & 0.0301 & 0.6665 & 0.9538 & 0.9551 & 0.9296 & 0.6017 & 0.3480 \\
& Ours     & 0.0096 & 0.6216 & 0.6111 & 0.6694 & 0.4898 & 0.7515 & 0.3053 \\
\hline
\multirow{2}{*}{\makecell[l]{Opposing-path /\\head-on-like}}
& Baseline & 0.0229 & 0.6667 & 0.8447 & 0.8690 & 0.8042 & 0.6177 & 0.3462 \\
& Ours     & 0.0082 & 0.6325 & 0.6460 & 0.7449 & 0.5218 & 0.8316 & 0.2871 \\
\hline
\multirow{2}{*}{\makecell[l]{Sideswipe / glancing}}
& Baseline & 0.0262 & 0.6744 & 0.9427 & 0.9355 & 0.9141 & 0.5894 & 0.3611 \\
& Ours     & 0.0090 & 0.6173 & 0.6332 & 0.7358 & 0.5638 & 0.7572 & 0.2944 \\
\hline
\end{tabular}%
\vspace{-4pt}
\end{table}

\section{Collision-Type Analysis}
\label{sec:supp-collision-types}

In the main paper we report aggregate results over the full CrashTwin test split. 
Here we provide a more fine-grained analysis by collision type and visualise representative examples. 
Our CARLA scripts instantiate seven directional intersection collision variants derived from the NHTSA pre-crash typology~\cite{najm2007precrash}. 
These script-level types differ in which arm the counterpart vehicle uses (oncoming, left, or right), whether the ego vehicle proceeds straight, turns left, or turns right, and whether the layout is a four-leg or a T-junction. 
For analysis we regroup these seven variants into three \emph{geometry-based} collision families according to the relative heading and impact angle between the two collision actors: near right-angle crossing impacts, opposing-path or head-on-like conflicts, and shallow-angle sideswipe or glancing collisions.

\subsection{Collision-Category Experimental Results}
\label{sec:supp-collision-types-quant}

We aggregate the seven script-level intersection types into three collision categories:
\begin{itemize}
    \item \textbf{Right-angle crossing.} 
    The lead accident vehicle proceeds straight and is struck from the side by a vehicle that crosses from the left or from the right at approximately a right angle on a four-leg intersection. 
    These correspond to the two ``straight versus crossing straight'' script variants and yield classic T-bone impacts.

    \item \textbf{Opposing-path / head-on-like conflicts.}
    The lead accident vehicle turns left and collides almost frontally with an oncoming straight vehicle or with a vehicle approaching from the left arm in such a way that the relative heading at impact is close to opposing. 
    This category groups two script variants (left-turn versus oncoming straight, and left-turn versus left-arm straight, on both four-leg and T-junction layouts) whose approach geometry is closer to a head-on or front-corner conflict than to a pure side impact.

    \item \textbf{Shallow-angle sideswipe / glancing collisions.}
    Both vehicles travel in roughly the same direction at impact and primarily graze along their sides while turning or merging. 
    This category includes the remaining three script variants (left-turn versus right-arm straight, right-turn versus oncoming left-turn, and right-turn versus left-arm straight, each on four-leg and T-junction layouts), where the two actors follow similar turning paths and only make shallow-angle contact.
\end{itemize}

For each collision category we recompute the main evaluation metrics on the corresponding subset of synthetic intersections and compare our post-trained model to the baseline world model~\cite{nvidia2025cosmosworldfoundationmodel}. 
Table~\ref{tab:collision-family-results} reports representative physics-grounded scores for each category.

Numerically, right-angle crossing and opposing-path categories show the most pronounced relative gains. 
For right-angle crossings, our model reduces the momentum residual from $0.95$ to $0.61$ and the energy-gain score from $0.93$ to $0.49$, while also lowering the warp error from $0.03$ to $0.01$ and increasing ID stability from $0.60$ to $0.75$. 
Opposing-path configurations exhibit similarly strong improvements: $J_p$ drops from $0.84$ to $0.65$, $J_E$ from $0.80$ to $0.52$, and $S_{\mathrm{ID}}$ increases from $0.62$ to $0.83$, accompanied by a reduction in appearance drift from $0.35$ to $0.29$. 
In these two categories, the approach geometry and impact configuration are tightly constrained by the intersection layout, so more accurate global reconstruction of approach and rebound motion directly translates into lower momentum and energy residuals and more stable identities.

Shallow-angle sideswipe and glancing collisions remain comparatively more challenging. 
Although our model still lowers $J_p$ from $0.94$ to $0.63$ and $J_E$ from $0.91$ to $0.56$, and improves $S_{\mathrm{ID}}$ from $0.59$ to $0.76$ with a noticeable reduction in appearance drift from $0.36$ to $0.29$, the residual momentum and energy scores remain higher than in right-angle crossings and the flow-divergence and appearance metrics show smaller absolute gains. 
This pattern reflects the intrinsic ambiguity of near-parallel trajectories and shallow contact angles, where small errors in lateral offset, depth, or mask alignment can change a light sideswipe into an apparent near-miss, and are therefore less easily corrected by improvements in the reconstruction pipeline.

\section{Additional Robustness Analyses}
\label{sec:supp-robustness}

This section reports additional robustness analyses used to validate the reliability of our reconstruction-based evaluator. These analyses are not used to redefine the benchmark metrics; they quantify calibration, error propagation, physical-model choices, window sensitivity, and reconstruction cost.

\subsection{Evaluator Calibration and Confidence}
\label{sec:supp-evaluator-calibration}

We compare scores computed from simulator metadata and from reconstructed outputs on the same synthetic videos in~\cref{tab:evaluator_calibration}. The gap measures calibration error, while the confidence intervals of benchmark means remain smaller than the median model gaps for the reported metrics.

\begin{table}[!t]
\centering
\scriptsize
\setlength{\tabcolsep}{3pt}
\renewcommand{\arraystretch}{1.10}
\vspace{-4pt}
\caption{\textbf{Evaluator calibration and score confidence.}
Simulator scores use simulator metadata, while reconstruction scores use the full reconstruction pipeline on the same synthetic videos.}
\label{tab:evaluator_calibration}
\vspace{-2pt}
\begin{tabular*}{\textwidth}{@{\extracolsep{\fill}}lcccc@{}}
\toprule
\textbf{Metric} &
\makecell{\textbf{Simulator}\\\textbf{metadata}} &
\makecell{\textbf{Reconstruction}\\\textbf{pipeline}} &
\makecell{\textbf{95\% confidence}\\\textbf{interval}} &
\makecell{\textbf{Median}\\\textbf{model gap}} \\
\midrule
$J_p$ & 0.0805 & 0.2721 & $\pm$0.0401 & 0.1034 \\
$J_H$ & 0.2067 & 0.3259 & $\pm$0.0370 & 0.0714 \\
$J_E$ & 0.0000 & 0.2051 & $\pm$0.0505 & 0.1214 \\
$S_{\mathrm{ID}}$ & 1.0000 & 0.8076 & $\pm$0.0209 & 0.0482 \\
$D_{\mathrm{ad}}$ & 0.2710 & 0.2810 & $\pm$0.0119 & 0.0236 \\
\bottomrule
\end{tabular*}
\vspace{-4pt}
\end{table}

The largest gaps appear in the collision-dependent physics scores, which is expected because they require metric-scale motion recovery from monocular videos. 
However, the confidence intervals of benchmark means remain below the median model gaps for all reported metrics, indicating that reconstruction error is not large enough to explain the main performance differences.

\subsection{Reconstruction Sensitivity Analysis}
\label{sec:supp-recon-sensitivity}

We perturb reconstructed 3D vehicle states to test how reconstruction errors propagate to physics scores. The $1\times$ perturbation scale is estimated from synthetic scenes by matching reconstructed vehicle frames to simulator annotations and measuring attribute differences, including center/depth, yaw/size, and tracking failures. We also scale the combined perturbation to $2\times$ and $5\times$ as stress tests. As shown in~\cref{tab:reconstruction_sensitivity}, empirical $1\times$ combined noise yields controlled changes, while larger perturbations produce larger deviations.

\begin{table}[t]
\centering
\scriptsize
\setlength{\tabcolsep}{3pt}
\renewcommand{\arraystretch}{1.10}
\vspace{-4pt}
\caption{\textbf{Reconstruction sensitivity analysis.}
Entries report mean absolute score changes after perturbing reconstructed 3D states.}
\label{tab:reconstruction_sensitivity}
\vspace{-2pt}
\begin{tabular*}{\textwidth}{@{\extracolsep{\fill}}lcccc@{}}
\toprule
\textbf{Perturbation source} & \textbf{Scale} & \textbf{Mean $|\Delta J_p|$} & \textbf{Mean $|\Delta J_H|$} & \textbf{Mean $|\Delta J_E|$} \\
\midrule
Center/depth & $1\times$ & 0.138 & 0.206 & 0.101 \\
Orient./size & $1\times$ & 0.000 & 0.087 & 0.000 \\
Tracking & $1\times$ & 0.055 & 0.073 & 0.062 \\
Combined & $1\times$ & 0.165 & 0.232 & 0.141 \\
Combined & $2\times$ & 0.235 & 0.291 & 0.214 \\
Combined & $5\times$ & 0.311 & 0.350 & 0.319 \\
\bottomrule
\end{tabular*}
\vspace{-4pt}
\end{table}

The perturbation study shows that position and depth errors mainly affect $J_p$ and $J_E$ through velocity estimation, while orientation and size errors primarily influence $J_H$. 
Tracking failures have smaller but non-negligible effects, and the monotonic increase from $1\times$ to $5\times$ confirms that the evaluator responds to reconstruction degradation in the expected direction.

\subsection{Friction-Aware Momentum Ablation}
\label{sec:supp-friction-ablation}

Exact tire-friction recovery from uncalibrated video is under-constrained because rolling or sliding friction mode and road condition are unobserved. We therefore test a conservative friction-aware correction by subtracting the maximum tire-road impulse $I_{\mathrm{fric}}=\sum_i \mu m_i g\Delta t$ from the momentum residual, using $\mu=0.8$ as a standard accident-reconstruction value~\cite{warner1983friction}. \Cref{tab:friction_ablation} shows that friction-aware residuals remain large and the ranking is nearly unchanged.

\begin{table}[t]
\centering
\scriptsize
\setlength{\tabcolsep}{3pt}
\renewcommand{\arraystretch}{1.10}
\vspace{-4pt}
\caption{\textbf{Friction-aware momentum ablation.}
We report raw and friction-aware $J_p$ values, together with the resulting rank changes.}
\label{tab:friction_ablation}
\vspace{-2pt}
\begin{tabular*}{\textwidth}{@{\extracolsep{\fill}}lcccccc@{}}
\toprule
\textbf{Model} &
\makecell{\textbf{Raw}\\$\boldsymbol{J_p}$} &
\makecell{\textbf{Friction-aware}\\$\boldsymbol{J_p}$} &
\makecell{\textbf{Change}\\$\boldsymbol{\Delta J_p}$} &
\makecell{\textbf{Raw}\\\textbf{rank}} &
\makecell{\textbf{Friction-aware}\\\textbf{rank}} &
\makecell{\textbf{Rank}\\\textbf{shift}} \\
\midrule
Cosmos-Predict2-14B & 0.6828 & 0.5661 & -0.1167 & 1 & 1 & 0 \\
Wan 2.1-14B & 0.8235 & 0.7756 & -0.0479 & 2 & 2 & 0 \\
Cosmos-Predict2-2B & 0.8890 & 0.8566 & -0.0323 & 3 & 4 & +1 \\
Wan 2.2-5B & 0.8899 & 0.8563 & -0.0336 & 4 & 3 & -1 \\
SkyReel-1.3B & 0.9566 & 0.9434 & -0.0132 & 5 & 5 & 0 \\
\bottomrule
\end{tabular*}
\vspace{-4pt}
\end{table}

The correction lowers $J_p$ for every model, but the remaining friction-aware residuals are still large. 
Thus, tire-road friction explains only a limited part of the momentum violations, and the original ranking is stable except for one swap between two very close models.

\subsection{Window-Size Ablation}
\label{sec:supp-window-ablation}

To validate our impact-window choice, we ablate $J_p$, $J_H$, and $J_E$ with $\pm3$, $\pm5$, and $\pm7$ frame windows, where $\pm X$ denotes $X$ frames before and after impact. Shorter windows stay closer to the impact and include less post-impact sliding, while longer windows include more post-impact motion and yield slightly larger residuals. We use $\pm5$ frames because it aligns with the short post-event Delta-V window in the U.S. federal event-data-recorder standard~\cite{nhtsa49cfr5637}. \Cref{tab:window_ablation} shows stable model ordering across these choices, supporting the middle $\pm5$ setting as a stable benchmark choice.

\begin{table}[t]
\centering
\scriptsize
\setlength{\tabcolsep}{5pt}
\renewcommand{\arraystretch}{1.10}
\vspace{-4pt}
\caption{\textbf{Window-size ablation.}
Ranks are reported in the $\pm3/\pm5/\pm7$ order.}
\label{tab:window_ablation}
\vspace{-2pt}
\begin{tabular}{@{}llcccc@{}}
\toprule
\textbf{Model} & \textbf{Metric} & \makecell{\textbf{$\pm3$}\\\textbf{frames}} & \makecell{\textbf{$\pm5$}\\\textbf{frames}} & \makecell{\textbf{$\pm7$}\\\textbf{frames}} & \textbf{Rank} \\
\midrule
\multirow{3}{*}{Cosmos-Predict2-14B} & $J_p$ & 0.6605 & 0.6856 & 0.7089 & 1/1/1 \\
& $J_H$ & 0.7088 & 0.7522 & 0.7677 & 1/1/1 \\
& $J_E$ & 0.6021 & 0.5967 & 0.5971 & 1/1/1 \\
\midrule
\multirow{3}{*}{Wan 2.1-14B} & $J_p$ & 0.7984 & 0.8138 & 0.8272 & 2/2/2 \\
& $J_H$ & 0.8182 & 0.8335 & 0.8448 & 2/2/2 \\
& $J_E$ & 0.7636 & 0.7688 & 0.7736 & 2/2/2 \\
\midrule
\multirow{3}{*}{Wan 2.2-5B} & $J_p$ & 0.8749 & 0.8853 & 0.8943 & 3/3/4 \\
& $J_H$ & 0.8811 & 0.8901 & 0.8995 & 3/3/4 \\
& $J_E$ & 0.8491 & 0.8567 & 0.8628 & 3/4/4 \\
\midrule
\multirow{3}{*}{Cosmos-Predict2-2B} & $J_p$ & 0.8754 & 0.8871 & 0.8940 & 4/4/3 \\
& $J_H$ & 0.8876 & 0.8917 & 0.8990 & 4/4/3 \\
& $J_E$ & 0.8524 & 0.8532 & 0.8533 & 4/3/3 \\
\midrule
\multirow{3}{*}{SkyReel-1.3B} & $J_p$ & 0.9380 & 0.9382 & 0.9391 & 5/5/5 \\
& $J_H$ & 0.9427 & 0.9461 & 0.9484 & 5/5/5 \\
& $J_E$ & 0.9227 & 0.9212 & 0.9138 & 5/5/5 \\
\bottomrule
\end{tabular}
\vspace{-4pt}
\end{table}

\subsection{Synthetic Oracle Reconstruction Validation and Runtime}
\label{sec:supp-oracle-runtime}

We further validate our reconstruction pipeline on synthetic clips with simulator ground truth. Beyond the trajectory ATE reported in the main paper, \cref{tab:synthetic_oracle_runtime} reports inter-vehicle distance, relative velocity, closing velocity, and per-vehicle speed errors. The results show that Kalman filtering brings the largest velocity gains, while metric depth correction gives the largest distance gain. The same table reports cumulative runtime per video on one NVIDIA RTX A5000 GPU, showing the accuracy-compute tradeoff of each added module.

\newsavebox{\syntheticOracleRuntimeTable}
\begin{table}[t]
\centering
\scriptsize
\setlength{\tabcolsep}{3pt}
\renewcommand{\arraystretch}{1.10}
\vspace{-4pt}
\caption{\textbf{Synthetic oracle reconstruction validation and runtime.}
Errors are measured against simulator metadata on matched synthetic clips; runtime is measured per video on one NVIDIA RTX A5000 GPU.}
\label{tab:synthetic_oracle_runtime}
\vspace{-2pt}
\sbox{\syntheticOracleRuntimeTable}{%
\begin{tabular}{lccccc}
\toprule
\multirow{2}{*}{\textbf{Configuration}} &
\multicolumn{4}{c}{\textbf{Mean absolute error}} &
\multirow{2}{*}{\makecell{\textbf{Runtime}\\\textbf{(s/video)}}} \\
\cmidrule(lr){2-5}
&
\makecell{\textbf{Inter-vehicle}\\\textbf{distance (m)}} &
\makecell{\textbf{Inter-vehicle}\\\textbf{velocity (m/s)}} &
\makecell{\textbf{Closing}\\\textbf{velocity (m/s)}} &
\makecell{\textbf{Vehicle}\\\textbf{speed (m/s)}} &
\\
\midrule
CenterTrack & 7.31 & 26.29 & 25.54 & 17.95 & 13.27 \\
+ Instance Relinking & 5.97 & 19.48 & 18.79 & 15.21 & 35.03 \\
+ Kalman Filtering & 5.73 & 6.18 & 6.10 & 4.95 & 40.89 \\
+ Metric Depth Correction & 3.09 & 3.92 & 3.98 & 4.09 & 167.55 \\
\bottomrule
\end{tabular}%
}
\ifdim\wd\syntheticOracleRuntimeTable>\textwidth
  \resizebox{\textwidth}{!}{\usebox{\syntheticOracleRuntimeTable}}%
\else
  \usebox{\syntheticOracleRuntimeTable}%
\fi
\vspace{-4pt}
\end{table}

\subsection{Synthetic and Real Split Evaluation}
\label{sec:supp-syn-real-split}

To separate simulator-controlled and in-the-wild evaluation settings, we report synthetic and real split results in~\cref{tab:syn-real-split}. 
The split results show that synthetic and real clips stress different aspects of the models: several models obtain lower visual warp errors on real clips, while physics scores remain challenging in both splits. 
This supports reporting the two settings separately rather than averaging away their distinct failure patterns.

\begin{table}[t]
\centering
\scriptsize
\setlength{\tabcolsep}{2pt}
\renewcommand{\arraystretch}{1.10}
\vspace{-4pt}
\caption{\textbf{Synthetic and real split evaluation.}
Open-source models are evaluated on 300 synthetic and 44 real clips, while proprietary models are evaluated on 100 synthetic and 16 real clips due to API limits. Lower values are better except for $S_{\mathrm{ID}}$.}
\label{tab:syn-real-split}
\vspace{-2pt}
\begin{tabular}{@{}ll|cc|ccc|cc@{}}
\multicolumn{2}{c|}{} &
\multicolumn{2}{c|}{\colorbox{orange!10}{\strut\makecell{Spatio-temporal\\Consistency}}} &
\multicolumn{3}{c|}{\colorbox{yellow!10}{\strut\makecell{Momentum \& Energy\\Conservation}}} &
\multicolumn{2}{c}{\colorbox{green!10}{\strut\makecell{World-dynamics\\Integrity}}} \\
\cline{1-9}
\textbf{Model} & \textbf{Subset} &
{\fontsize{6.5pt}{7.0pt}\selectfont \makecell{\rule{0pt}{2.6ex}Warp\\Error}} &
{\fontsize{6.5pt}{7.0pt}\selectfont \makecell{\rule{0pt}{2.6ex}Flow\\Div.}} &
{\fontsize{6.5pt}{7.0pt}\selectfont \makecell{\rule{0pt}{2.6ex}Momentum\\Residual}} &
{\fontsize{6.5pt}{7.0pt}\selectfont \makecell{\rule{0pt}{2.6ex}Angular\\Residual}} &
{\fontsize{6.5pt}{7.0pt}\selectfont \makecell{\rule{0pt}{2.6ex}Energy\\Gain}} &
{\fontsize{6.5pt}{7.0pt}\selectfont \makecell{\rule{0pt}{2.6ex}Instance\\Stability}} &
{\fontsize{6.5pt}{7.0pt}\selectfont \makecell{\rule{0pt}{2.6ex}App.\\Drift}} \\
\cline{3-9}
& &
{\fontsize{6.5pt}{7.0pt}\selectfont $E_{\mathrm{warp}} \downarrow$} &
{\fontsize{6.5pt}{7.0pt}\selectfont $E_{\mathrm{div}} \downarrow$} &
{\fontsize{6.5pt}{7.0pt}\selectfont $J_p \downarrow$} &
{\fontsize{6.5pt}{7.0pt}\selectfont $J_H \downarrow$} &
{\fontsize{6.5pt}{7.0pt}\selectfont $J_E \downarrow$} &
{\fontsize{6.5pt}{7.0pt}\selectfont $S_{\mathrm{ID}} \uparrow$} &
{\fontsize{6.5pt}{7.0pt}\selectfont $D_{\mathrm{ad}} \downarrow$} \\
\hline
\multirow{2}{*}{SkyReel-1.3B} & Synthetic & 0.0229 & 0.6175 & 0.9564 & 0.9640 & 0.9440 & 0.6637 & 0.3555 \\
& Real & 0.0214 & 0.5606 & 0.9579 & 0.9545 & 0.9577 & 0.6850 & 0.3845 \\
\hline
\multirow{2}{*}{Wan 2.1-14B} & Synthetic & 0.0188 & 0.6287 & 0.8306 & 0.8498 & 0.7866 & 0.6792 & 0.3118 \\
& Real & 0.0104 & 0.6582 & 0.7524 & 0.8466 & 0.7848 & 0.6528 & 0.3108 \\
\hline
\multirow{2}{*}{Wan 2.2-5B} & Synthetic & 0.0155 & 0.5939 & 0.9073 & 0.9087 & 0.8834 & 0.7229 & 0.3132 \\
& Real & 0.0071 & 0.6097 & 0.7453 & 0.8210 & 0.7388 & 0.7422 & 0.2951 \\
\hline
\multirow{2}{*}{Cosmos-Predict2-2B} & Synthetic & 0.0263 & 0.6690 & 0.9111 & 0.9180 & 0.8796 & 0.6036 & 0.3514 \\
& Real & 0.0081 & 0.7156 & 0.6993 & 0.7373 & 0.7146 & 0.6765 & 0.3098 \\
\hline
\multirow{2}{*}{Cosmos-Predict2-14B} & Synthetic & 0.0124 & 0.7195 & 0.6855 & 0.7629 & 0.5950 & 0.6749 & 0.3333 \\
& Real & 0.0071 & 0.7078 & 0.6583 & 0.7630 & 0.6741 & 0.6651 & 0.3285 \\
\hline
\multirow{2}{*}{Google Veo 3.1} & Synthetic & 0.0105 & 0.6184 & 0.7740 & 0.7862 & 0.7379 & 0.7367 & 0.3073 \\
& Real & 0.0044 & 0.6309 & 0.7774 & 0.9004 & 0.8000 & 0.6330 & 0.3090 \\
\hline
\multirow{2}{*}{Hailuo 2.3} & Synthetic & 0.0166 & 0.6130 & 0.7725 & 0.7837 & 0.7209 & 0.6007 & 0.3002 \\
& Real & 0.0053 & 0.6225 & 0.7112 & 0.7648 & 0.7797 & 0.7428 & 0.2739 \\
\hline
\end{tabular}%
\vspace{-4pt}
\end{table}

\begin{figure}[t]
    \centering
    \includegraphics[width=0.9\textwidth]{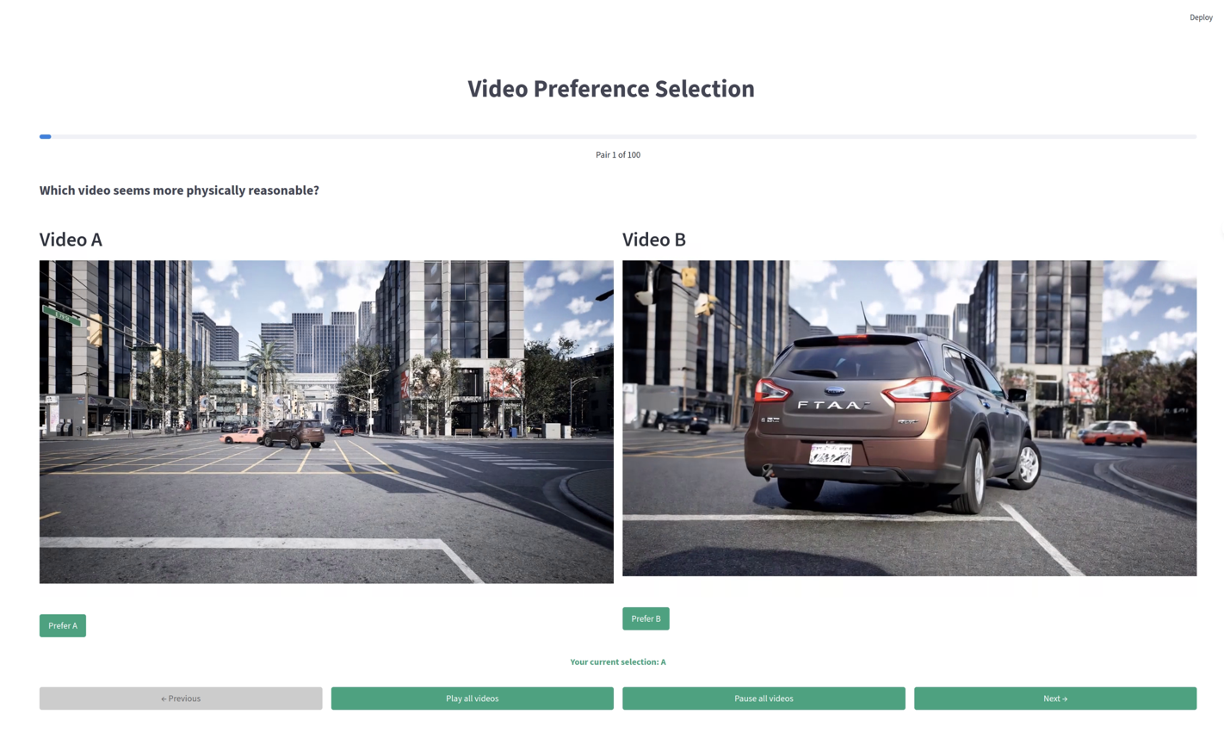}
    \caption{User interface for the human 2AFC study. Each page displays two rollouts from the same scenario (left: A, right: B) with anonymized model identities, together with the question ``Which video seems more physically reasonable?'' and radio buttons for selecting A or B. Annotators can replay both videos before answering, and a progress bar at the top indicates completion status of the session.}
    \label{fig:human_eval_interface}
\end{figure}

\section{Human Evaluation.}
\label{sec:humaneval}
To better quantify how our physics-based metrics relate to perceptual judgments, we conduct a two-alternative forced-choice (2AFC)~\cite{green1966signal} user study, following recent work~\cite{cong2025wfm-tts} on human evaluation for video generation. We recruit 10 graduate participants, each of whom annotates around 100 queries, yielding a total of 1{,}000 pairwise responses. For every query, we present two rollouts from the same initial scenario side by side, anonymize model identities, and ask annotators: ``Which video seems more physically reasonable?''. The web interface used in this study is illustrated in~\cref{fig:human_eval_interface}.

Let $T$ be the total number of annotated pairs, and let $(A_t,B_t,y_t)$ denote the two rollouts and the human choice for pair $t \in \{1,\dots,T\}$, where $y_t \in \{A,B\}$. For a given metric $k$, which induces a deterministic preference $m_k(A_t,B_t) \in \{A,B\}$ on each pair, we define its 2AFC accuracy as
\begin{equation}
    \mathrm{Acc}^{(k)}_{\text{2AFC}}
    = \frac{1}{T} \sum_{t=1}^{T}
      \mathbf{1}\bigl[m_k(A_t,B_t) = y_t\bigr],
\end{equation}
where $\mathbf{1}[\cdot]$ is the indicator function. This is the standard 2AFC proportion-correct measure: it directly reflects how often the metric selects the same rollout as human annotators.

In all our analyses, we simply use $\mathrm{Acc}^{(k)}_{\text{2AFC}}$ as the
human-alignment score for metric $k$, and report these scores for both
physics-based and appearance-based metrics to compare how well each family
agrees with human judgements of physical realism.

\section{Post Training Details}
\label{sec:PostTrainingInference}

Post training is performed on the Cosmos Predict2 2B video to world model~\cite{nvidia2025cosmosworldfoundationmodel} with LoRA~\cite{hu2022lora} on 8 $\times$ A100 GPUs. LoRA is configured with rank 16 and alpha 16 and is injected into the attention projections together with the 2 MLP layers inside each transformer block. The post-training corpus consists of collision-focused video clips with at least 93 frames.
Clips shorter than 93 frames are padded by repeating the final frame until reaching a length of 93, after which all clips are resized to $704 \times 1280$. Training uses batch size 1 under fully sharded data parallelism with context parallel size 4. Optimization relies on Fused AdamW~\cite{loshchilov2019decoupled} with a learning rate of $2^{-9.5}$ together with a linear schedule over 10000 iterations. During inference the LoRA weights are merged into the base model and the standard video to world generation pipeline is applied.

\subsubsection*{Acknowledgements.}
This work was supported in part by GPU hardware and cloud credits provided to the Texas A\&M University PHAI Lab and TACO Lab through the NVIDIA Academic Grant Program. We also acknowledge computational resources provided by the Texas A\&M VISION GPU Cluster. Z. Fan was supported in part by research gifts from Meta and the US AI Glasses Impact Grants program.
J. Hong is supported by NAIRR250526. Z. Wang is supported by DARPA ANSR (RTX CW2231110), DARPA TIAMAT (HR0011-24-9-0431), ARL StAmant (W911NF-23-S-0001), as well as the NSF AI Institute for Foundations of Machine Learning (IFML).

\bibliographystyle{assets/plainnat}
\bibliography{main}

\end{document}